\PassOptionsToPackage{table,xcdraw}{xcolor}
\documentclass[pdflatex,iicol]{sn-jnl}

\usepackage{graphicx}%
\usepackage{multirow}%
\usepackage{amsmath,amssymb,amsfonts}%
\usepackage{amsthm}%

\usepackage{mathrsfs}%
\usepackage[title]{appendix}%
\usepackage{booktabs}%
\usepackage{fix-cm}
\usepackage[most]{tcolorbox}
\usepackage{lipsum} % for dummy text
\usepackage{caption}
\usepackage{subcaption}
\usepackage{dirtree}
\usepackage{pifont}
\numberwithin{equation}{section}
\usepackage{times}
\usepackage{helvet}
\usepackage{colortbl}
\usepackage{footnote}

\usepackage{amsmath,amsfonts,bm}

\def\eqref#1{equation~\ref{#1}}
\def\1{\bm{1}}

\DeclareMathAlphabet{\mathsfit}{\encodingdefault}{\sfdefault}{m}{sl}
\SetMathAlphabet{\mathsfit}{bold}{\encodingdefault}{\sfdefault}{bx}{n}

\usepackage{booktabs}
\usepackage{graphicx}
\usepackage[table]{xcolor}

\definecolor{bettergreen}{RGB}{0,120,60}
\definecolor{worsered}{RGB}{180,35,35}

\newcommand{\better}[1]{\textcolor{bettergreen}{#1}}
\newcommand{\worse}[1]{\textcolor{worsered}{#1}}

\newcommand{\hneg}[2]{#1\,{\scriptsize\worse{(-#2)}}}

\newcommand{\lpos}[2]{#1\,{\scriptsize\better{(+#2)}}}
\newcommand{\lneg}[2]{#1\,{\scriptsize\worse{(-#2)}}}

\newcommand{\Tref}[1]{Tab.~\ref{#1}}
\newcommand{\Eref}[1]{Eq.~(\ref{#1})}
\newcommand{\Fref}[1]{Fig.~\ref{#1}}
\newcommand{\Sref}[1]{Sec.~\ref{#1}}

\newcommand{\ie}{\textit{i}.\textit{e}.}
\newcommand{\eg}{\textit{e}.\textit{g}.}

\newcommand{\bench}{\textsc{VagueBench}}

\theoremstyle{thmstyleone}%
\theoremstyle{thmstyletwo}%

\theoremstyle{thmstylethree}%

\begin{document}

\title[Article Title]{\bench: Benchmarking the Robustness of Embodied Agents on Ambient Context Illusions}

%%=============================================================%%
%% GivenName	-> \fnm{Joergen W.}
%% Particle	-> \spfx{van der} -> surname prefix
%% FamilyName	-> \sur{Ploeg}
%% Suffix	-> \sfx{IV}
%% \author*[1,2]{\fnm{Joergen W.} \spfx{van der} \sur{Ploeg} 
%%  \sfx{IV}}\email{iauthor@gmail.com}
%%=============================================================%%

\author[1]{\fnm{Yang} \sur{Qu}}
% \email{qyang@buaa.edu.cn}
\equalcont{These authors contributed equally to this work.}

\author[1,5]{\fnm{Qiang} \sur{Fu}}
% \email{fuq@buaa.edu.cn}
\equalcont{These authors contributed equally to this work.}

\author[1]{\fnm{Zonghao} \sur{Ying}}
% \email{nyale1024@gmail.com}

\author[4]{\fnm{Jiayi} \sur{Zhao}}
% \email{23373278@buaa.edu.cn}

\author[1]{\fnm{Yaoyuan} \sur{Zhang}}
% \email{yaoyzhang@buaa.edu.cn}

\author[1]{\fnm{Jinyang} \sur{Guo}}
% \email{guojinyang@buaa.edu.cn}

\author[1]{\fnm{Jian} \sur{Yang}}
% \email{jiaya@buaa.edu.cn}

\author*[1]{\fnm{Aishan} \sur{Liu}}\email{liuaishan@buaa.edu.cn}

\author[5]{\fnm{Yingxun} \sur{Wang}}
% \email{wangyx@buaa.edu.cn}

\author[1,2,3]{\fnm{Xianglong} \sur{Liu}}
% \email{xlliu@buaa.edu.cn}

\affil[1]{\orgname{SKLCCSE, Beihang University}, \orgaddress{\country{China}}}

% \affil[2]{\orgname{Henan University of Science and Technology}, \orgaddress{\country{China}}}

\affil[2]{\orgname{Zhongguancun Laboratory}, \orgaddress{\country{China}}}

\affil[3]{\orgname{Institute of Dataspace}, \orgaddress{\country{China}}}

\affil[4]{\orgname{School of Artificial Intelligence, Beihang University}, \orgaddress{\country{China}}}

\affil[5]{\orgname{School of Automation Science and Electrical Engineering, Beihang University}, \orgaddress{\country{China}}}

%%==================================%%
%% Sample for unstructured abstract %%
%%==================================%%

\abstract{
Embodied agents are increasingly deployed in open-world environments, yet current robustness evaluations remain biased toward hazardous instructions and jailbreak vulnerabilities. A more fundamental issue is largely overlooked: agents are rarely tested under semantically sufficient but contextually minimal instructions (\ie, common in real-world human communication), where auxiliary contextual cues are absent. For the first, we identify a novel phenomenon termed \emph{Ambient Context Illusion}, where agents fail to execute tasks correctly when non-essential contextual details are omitted, despite unchanged task semantics. To systematically study this, we establish \bench, a benchmark built upon the AI2-THOR interactive simulation environment for the evaluation of context dependency in embodied agents. We systematically consider key semantic components of instructions (entities, procedures, constraints, and prior knowledge) and derive a unified taxonomy with four categories and nine subtypes to reduce auxiliary information. \bench\ contains 1,303 instruction pairs across diverse indoor environments, constructed via a human–agent co-generation framework that integrates semantic redundancy estimation, taxonomy-guided transformation, and rigorous semantic-invariance verification. Through extensive evaluation on 6 state-of-the-art embodied agents, we observe significant degradation on \bench\ in both task performance and executability (relatively -19.65\% in task success rate and relatively +54.10\% in empty plan rate) revealing a strong reliance on auxiliary linguistic signals rather than core task semantics. Furthermore, we demonstrate that this phenomenon persists in real-world robotic systems, where it can lead to incorrect decisions and potential safety risks. To mitigate this issue, we propose a lightweight, model-agnostic method Context Suppression Constrain that enforces goal-centric semantic grounding, achieving significant robustness improvements ($\times$3.36). Code, data, and demos are available at \url{https://github.com/stardreaming/VagueBench}.
}


%%\pacs[JEL Classification]{D8, H51}

%%\pacs[MSC Classification]{35A01, 65L10, 65L12, 65L20, 65L70}
\maketitle
\section{Introduction}

Embodied agents, which integrate perception, language understanding, and physical interaction, are increasingly deployed in open-world environments such as household robotics and human–robot collaboration \cite{driess2023palm,zitkovich2023rt,xu2024surveyroboticsfoundationmodels,zhang2024empowering,cao2019ghostar}. Powered by vision–language models (VLMs), these agents can interpret natural language instructions and execute complex, long-horizon tasks in partially observable environments. As such systems move closer to real-world deployment, evaluating their robustness under realistic conditions has become a critical research problem.

However, current robustness evaluation paradigms for VLM-driven embodied agents remain fundamentally limited. Existing benchmarks predominantly focus on adversarial settings, such as hazardous instructions, jailbreak attacks, or safety-critical edge cases \cite{liu2020spatiotemporal,zhang2024badrobot,liu2025compromising,ying2025agentsafe,zhang2025realistic,yin2024safeagentbench}. While important, these evaluations overlook a more pervasive and realistic scenario: agents are rarely tested under semantically sufficient but contextually minimal instructions, where auxiliary contextual cues are absent. However, such instructions are ubiquitous in real-world human communication, where users naturally omit redundant details. For example, 

\begin{figure}[!t]
\centering
\includegraphics[width=\linewidth,trim=120 150 320 210, clip]{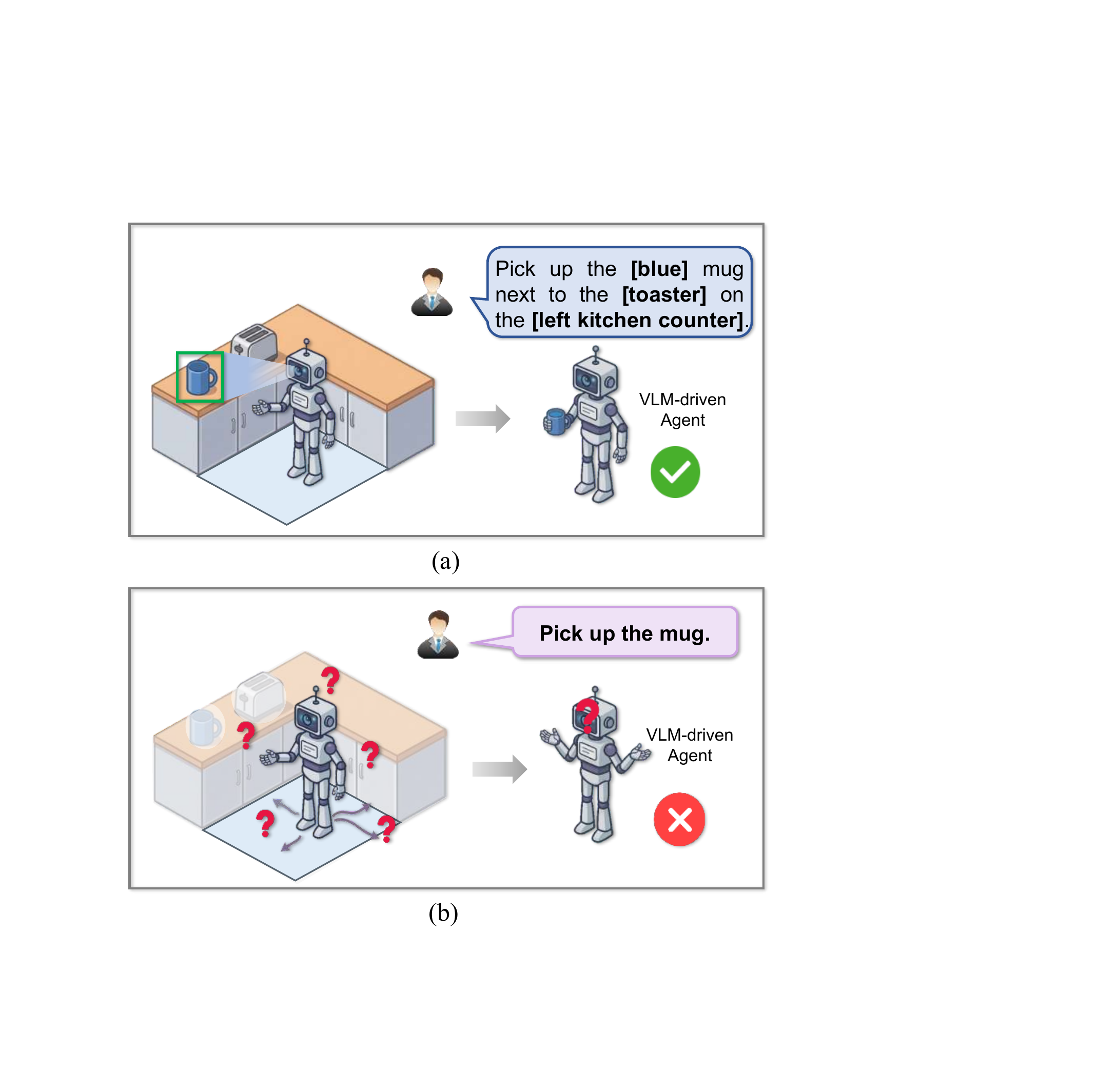}
\caption{Illustration of the Ambient Context Illusion. (a) the normal case, where the agent understand the instructions and finish the task. (b) the ambient context illusion case, where agents fail to execute tasks correctly when non-essential contextual details are omitted.}
\label{fig:illusion_example}
\end{figure}

This gap raises a critical question: do current agents truly ground task semantics under minimal instructions, or do they implicitly rely on redundant contextual cues
In this work, we, for the first time, formalize this phenomenon as Ambient Context Illusion (\Fref{fig:illusion_example}), defined as the tendency of embodied agents to produce incorrect behaviors when non-essential contextual details are omitted, despite the task remaining well-defined and executable. Unlike classical hallucination or ambiguity \cite{ji2023survey,anderson2018vision,ho2024novo,ho2025review}, this illusion arises from an over-reliance on ambient linguistic signals (information that is helpful but not necessary), highlighting a fundamental weakness in semantic grounding\cite{chen2025mllms,liu2025roboview}. To systematically investigate this, we establish \bench, a benchmark designed to evaluate context dependency in embodied agents. \bench\ is built upon the AI2-THOR \cite{kolve2017ai2} simulation environment and constructs paired instructions via semantically-preserving compression, which removes auxiliary contextual information while keeping task semantics unchanged. We further propose a unified taxonomy that captures multiple dimensions of contextual reduction, enabling fine-grained analysis of agent behavior under minimal instructions. The resulting dataset contains 1,303 high-quality instruction pairs across diverse indoor environments.

We conduct extensive experiments on six state-of-the-art embodied agents under multiple workflow paradigms (e.g., Qwen2.5-VL-72B-Instruct \cite{bai2025qwen25vltechnicalreport}, Qwen2-VL-72B-Instruct \cite{wang2024qwen2}, InternVL-3-72B \cite{zhu2025internvl3}, GPT-4o \cite{hurst2024gpt4o}, ReAct \cite{yao2022react} and ProgPrompt \cite{singh2023progprompt}). The results reveal consistent and substantial performance degradation under context-reduced instructions (\eg, an average relative drop of 19.65\% in task success rate and an average relative increase of 54.10\% in empty plan rate). These findings demonstrate that current agents heavily depend on auxiliary linguistic cues rather than robust semantic grounding. Moreover, we show that this phenomenon persists in real-world robotic systems, where it can lead to incorrect decisions and potential safety risks\cite{liu2025agentsafe,liang2025safemobile}.

To mitigate this issue, we propose a lightweight, model-agnostic method termed Context Suppression Constraint, which enforces goal-centric semantic grounding and discourages reliance on implicit contextual assumptions. Despite its simplicity, this approach significantly improves robustness, increasing by 2.36 times under context-reduced settings. We hope this work draws attention to this overlooked failure mode and encourages the development of evaluation protocols and agent designs that prioritize robust semantic grounding under realistic human instructions. Our contributions are:
\begin{itemize}
\item We identify \textbf{Ambient Context Illusion}, a novel failure mode revealing the over-reliance of embodied agents on non-essential contextual information under semantically valid instructions.
\item We introduce \bench, a benchmark for evaluating context dependency via semantically-preserving instruction compression, supported by a unified taxonomy.
\item We conduct comprehensive evaluations across multiple models and workflows, demonstrating significant performance degradation and uncovering systematic failure patterns.
\end{itemize}

\section{Related Work}
VLM-driven embodied agents map natural language instructions to sequential actions in interactive environments, enabling applications in robotics, virtual assistants, and simulation \cite{zhang2023building,batra2020objectnav,wang2023voyager,zhang2025up}.
As these systems move toward real-world deployment, robust and reliable evaluation benchmarks become essential for assessing their behavior under realistic conditions \cite{ying2026safebench,xiao2025detoxifying,ying2025jailbreak,liu2025pre}.

\textbf{General Evaluation Benchmarks.} Recent progress in embodied agents has been driven by simulation platforms and language-conditioned task datasets. Environments such as AI2-THOR \cite{kolve2017ai2} and Habitat \cite{savva2019habitat} provide interactive 3D scenes for evaluating perception and action, while datasets like ALFRED \cite{shridhar2020alfred} introduce long-horizon instruction-following tasks. Subsequent benchmarks, including ALFWorld \cite{shridhar2020alfworld} and EmbodiedBench \cite{yang2025embodiedbench}, further extend evaluation to scalable and compositional settings.
However, these benchmarks typically assume richly specified instructions with detailed object attributes, spatial relations, and contextual cues, which simplify grounding and reduce uncertainty.

%In the context of \emph{general evaluation benchmarks} for embodied agents, recent advances have been largely driven by simulation platforms and language-conditioned task datasets. Environments such as AI2-THOR \cite{kolve2017ai2} and Habitat \cite{savva2019habitat} provide interactive 3D scenes for evaluating perception and action, while datasets like ALFRED \cite{shridhar2020alfred} introduce long-horizon instruction-following tasks that require grounding natural language into sequential actions. Subsequent benchmarks, including ALFWorld \cite{shridhar2020alfworld} and EmbodiedBench \cite{yang2025embodiedbench}, further extend evaluation to more scalable and compositional settings.  These benchmarks typically assume richly specified instructions containing detailed object attributes, spatial relations, and contextual cues, which provide strong grounding signals and simplify decision-making.

\textbf{Robustness and Safety Evaluation.} Recent work has increasingly examined the robustness of embodied agents under realistic conditions\cite{ying2026safebench,liang2024object,zhang2024visual}. Benchmarks such as ReliabilityBench \cite{gupta2026reliabilitybench} and AgentNoiseBench \cite{wang2026agentnoisebench} evaluate sensitivity to noisy instructions, ambiguity, and distribution shifts, while datasets like AmbiK \cite{ivanova2025ambik} focus on resolving underspecified or conflicting language.
In parallel, safety-oriented benchmarks such as AgentSafe \cite{ying2025agentsafe,liu2025agentsafe} and SafeAgentBench \cite{yin2024safeagentbench} assess agent behavior under hazardous or adversarial conditions \cite{liu2024compromising,wang2025manipulating,ying2024jailbreak,lu2025adversarial,zeng2025safesteer}. Other studies highlight hallucination issues arising from imperfect cross-modal grounding \cite{chakraborty2025heal}.
Overall, these works primarily characterize robustness as resilience to external perturbations or ill-posed inputs.

\textbf{Comparison.}
Existing studies primarily attribute failures to external perturbations (\eg, noise, distribution shifts) or imperfect task specifications. In contrast, our work investigates a more subtle yet realistic setting where instructions remain \emph{clean, semantically correct, and fully executable}, but are simplified by removing non-essential contextual information. Our work differs from prior research in three key aspects. \ding{182} Problem formulation. Prior work focuses on robustness under perturbed or ill-posed inputs, whereas we study robustness under \emph{semantically equivalent but context-minimal instructions}, reflecting natural human communication. \ding{183} Input assumption. Existing benchmarks introduce noise or ambiguity, making failures difficult to disentangle. In contrast, our setting preserves semantic completeness and executability, ensuring that performance degradation stems from context dependency rather than missing or corrupted information. \ding{184} Evaluation focus. Prior evaluations emphasize safety violations or task failure under perturbations, while we reveal a latent failure mode (\ie, Ambient Context Illusion), characterized by agents’ reliance on auxiliary contextual cues instead of robust semantic grounding \cite{liang2024badclip,liang2025vl,liang2025revisiting,liang2024unlearning,kuang2024adversarial,xun2025robust}.

\section{Preliminaries and Backgrounds}

\subsection{VLM-Driven Embodied Task}

We consider language-conditioned embodied tasks in partially observable environments, where an agent interacts with the environment to accomplish a task specified by a natural language instruction. 
Modern VLM-driven agents typically operate in an iterative planner-interaction loop, where high-level semantic decisions are interleaved with environment feedback.

In particular, let $\bm{I}$ denote the instruction, $\bm{o}_t$ the observation at step $t$, and $\bm{h}_t$ the past interaction history. The agent produces a semantic plan
\begin{equation}
\bm{z}_t = f_{\Phi}(\bm{I}, \bm{o}_t, \bm{h}_t),
\label{eq:planner}
\end{equation}
where $f_{\Phi}$ denotes the VLM-based planner, and $\bm{z}_t$ represents an intermediate task interpretation (\eg, subgoal or action plan).

The plan $\bm{z}_t$ is then executed through interaction with the environment, inducing state transition and history update:
\begin{equation}
(\bm{o}_{t+1}, \bm{h}_{t+1}) = E(\bm{z}_t, \bm{o}_t, \bm{h}_t),
\label{eq:transition}
\end{equation}
where $E(\cdot)$ abstracts environment dynamics together with execution and feedback. 
The interaction process induces an execution trajectory $\bm{\tau}$, which consists of the sequence of observations and intermediate plans generated during interaction. 
The trajectory is implicitly determined by the planner $f_{\Phi}$ and the transition operator $E(\cdot)$.

This formulation decouples semantic planning from environment interaction, and subsumes a broad class of embodied agents, including single-step policies, multi-step planners, and reasoning–acting frameworks.

\subsection{Problem Definition}

In real-world human communication, instructions are often \emph{concise and contextually incomplete}, yet remain fully understandable and executable \cite{dogan2025model,bila2026contextual}. 
Humans naturally rely on shared context and commonsense reasoning to infer omitted details, adhering to principles such as parsimony (\eg, Occam’s razor \cite{blumer1987occam}), where non-essential information is omitted without affecting task interpretation. 
As a result, semantically sufficient instructions do not require exhaustive specification of object attributes, spatial relations, or procedural details. This observation raises a fundamental question for embodied agents: can agents maintain consistent behavior when auxiliary contextual information is removed, while task semantics remain unchanged?

In this work, we formalize the failure of this capability as \textbf{Ambient Context Illusion}. Specifically, given two instructions $\bm{I}$ and $\tilde{\bm{I}}$ that are semantically equivalent and both executable in the environment, an agent exhibits Ambient Context Illusion if it produces divergent semantic interpretations or behaviors when executing these instructions.

To formalize this, we consider a semantically-preserving transformation as
\begin{equation}
\tilde{\bm{I}} = T(\bm{I}),
\end{equation}
where $T(\cdot)$ removes auxiliary contextual cues while preserving the core task semantics. 
We refer to the transformed instruction $\tilde{\bm{I}}$ as a \emph{vague instruction}, which is characterized by reduced contextual information while maintaining semantic equivalence and executability.

We ensure semantic equivalence via a human-centric verification function:
\begin{equation}
\mathcal{V}_{human}(\bm{I}, \tilde{\bm{I}}) = 
\begin{cases} 
1, & \text{if core semantic matched,} \\
0, & \text{otherwise.}
\end{cases}
\end{equation}
Specifically, $\mathcal{V}_{human}(\bm{I}, \tilde{\bm{I}}) = 1$ if and only if independent human annotators unanimously verify that the original instruction $\bm{I}$ and the vague instruction $\tilde{\bm{I}}$ invoke the exact same core task intent, and that $\tilde{\bm{I}}$ remains unambiguously executable within the given environment. (The detailed verification protocol is presented in \Sref{4.3}).

Despite this semantic equivalence, the agent may produce different semantic plans:
\begin{equation}
f_{\Phi}(\bm{I}, \bm{o}_t, \bm{h}_t) \neq f_{\Phi}(\tilde{\bm{I}}, \bm{o}_t, \bm{h}_t),
\end{equation}
which further leads to divergent interaction trajectories and outcomes.

At the behavioral level, we quantify this effect as a performance gap:
\begin{equation}
\Delta =
\mathbb{E}_{\bm{\tau}}
\left[
R(\bm{\tau} \mid \bm{I}) - R(\bm{\tau} \mid \tilde{\bm{I}})
\right],
\end{equation}
where $\bm{\tau}$ denotes the execution trajectory and $R(\cdot)$ measures task performance.

A significant $\Delta>0$ indicates that the agent fails to maintain consistent behavior under semantically equivalent instructions, revealing a reliance on auxiliary contextual cues rather than robust semantic grounding. Importantly, this phenomenon is distinct from ambiguity or perceptual uncertainty: both $\bm{I}$ and $\tilde{\bm{I}}$ are fully executable and unambiguous. 
Thus, Ambient Context Illusion reflects a fundamental limitation in how agents interpret task semantics under context reduction.

\section{Benchmark Design}

\begin{figure*}[!t]
\centering
\includegraphics[width=\textwidth,trim=40 260 170 260, clip]{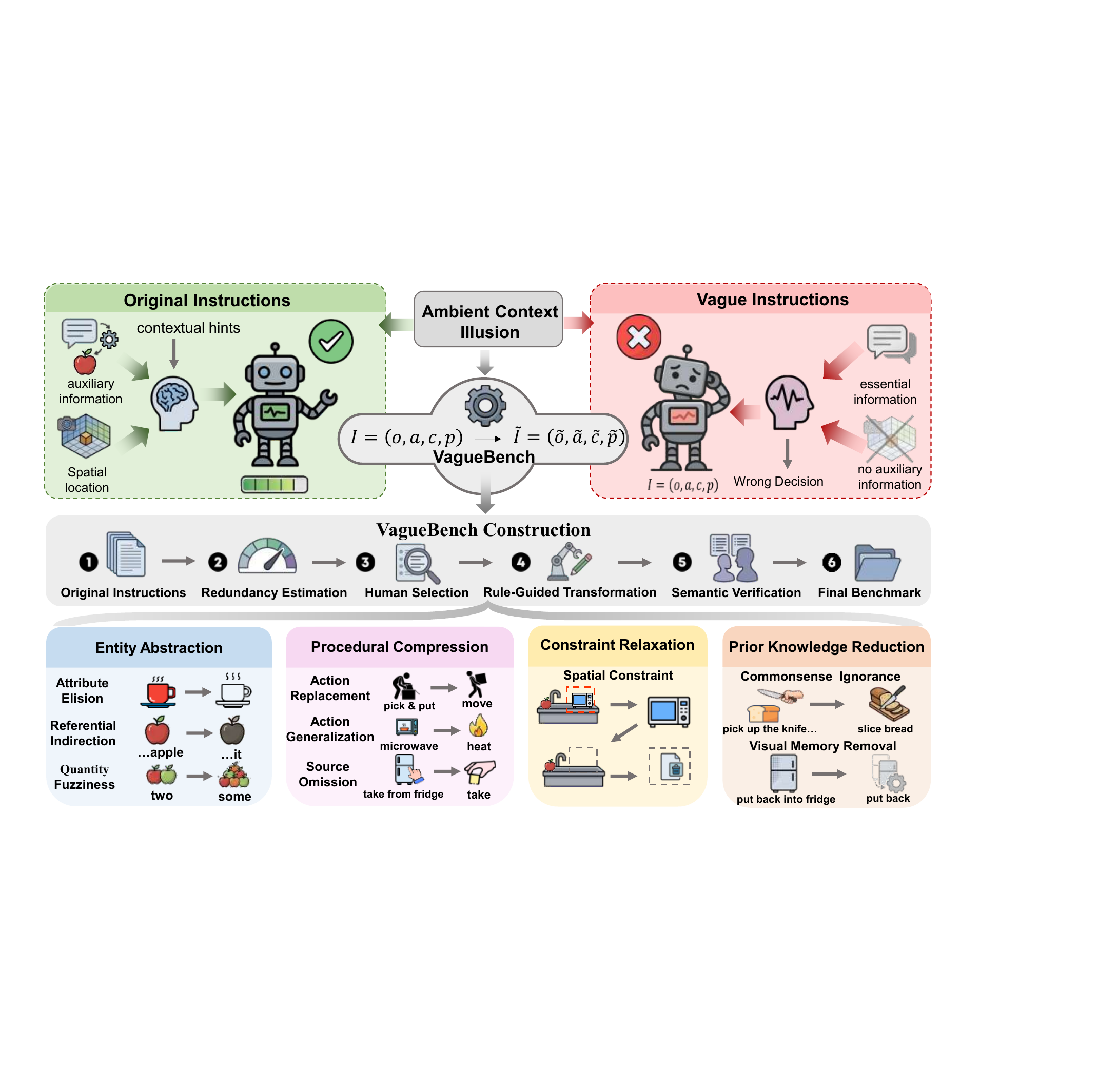}
\caption{Overview of \bench\ and the Ambient Context Illusion paradigm, which illustrates how embodied agents rely on auxiliary contextual cues in original instructions and presents the \bench\ construction pipeline and a taxonomy of instruction transformation.}
\label{fig:total_frame}
\end{figure*}

As shown in \Fref{fig:total_frame}, we introduce \bench\ and a human–agent co-generation pipeline, to systematically study the effect of redundant contextual cues on embodied agents.
To formalize the reduction of these contextual cues, and following prior works in semantic parsing and instruction grounding for embodied tasks \cite{shin2025socratic,lu2023thinkbot,nyga2018grounding,yang2026imprif}, we instantiate auxiliary information removal as structured transformations over different semantic dimensions of the instruction. Specifically, we conceptualize the semantic elements of an instruction along four key dimensions: entity descriptions ($\bm{e}$), procedural specifications ($\bm{a}$), task constraints ($\bm{c}$), and prior knowledge dependencies ($\bm{p}$).

It is worth noting that natural language instructions are inherently flexible; thus, not all instructions explicitly contain all four dimensions. Certain components may be implicitly embedded within the linguistic context or omitted entirely. Therefore, rather than defining an instruction as a rigid tuple, we represent it as a composition of these potential semantic features:
\begin{equation}
\bm{I} \sim \{ \bm{e}, \bm{a}, \bm{c}, \bm{p}\}
\label{eq:instruction_factorization}
\end{equation}
Accordingly, we systematically design a component-wise transformation:
\begin{equation}
\tilde{\bm{I}} = T(\bm{I}) \sim \{\tilde{\bm{e}}, \tilde{\bm{a}}, \tilde{\bm{c}}, \tilde{\bm{p}}\},
\label{eq:transformation}
\end{equation}
where each available component is selectively simplified while preserving the underlying task semantics.
% We represent an instruction as:
% \begin{equation}
% I = (e, a, c, p),
% \label{eq:instruction_factorization}
% \end{equation}
% where $e$, $a$, $c$, and $p$ denote entity descriptions, procedural specifications, task constraints, and prior knowledge dependencies, respectively.
% We systematically design a component-wise transformation $\tilde{I} = T(I) = (\tilde{e}, \tilde{a}, \tilde{c}, \tilde{p})$, where each component is selectively simplified while preserving the underlying task semantics.

\subsection{\bench\ Taxonomy}\label{4.1}

We organize \bench\ into four categories, each corresponding to a structured transformation over a specific component of \Eref{eq:instruction_factorization}. Each category further consists of several subcategories that instantiate concrete transformation patterns.

\ding{182} \textbf{Entity Abstraction.}
This category corresponds to abstracting the part of specially detailed information of entity in instructions. 
Given an entity description $o$, we remove the auxiliary information of it like: 
\begin{equation}
\tilde{\bm{e}} = T_e(\bm{e}), \quad \tilde{\bm{e}} \preceq \bm{e},
\end{equation}
where $\preceq$ denotes a partial order over semantic completeness. 
The transformation preserves referential correctness to ensure the only target while removing auxiliary descriptive cues, thereby weakening explicit grounding signals. 
Entity Abstraction includes the three subcategories: \emph{Attribute Elision}, \emph{Referential Indirection}, and \emph{Quantity Fuzziness}.
In practice, this includes \emph{Attribute Reduction}, Referential Indirection, and Quantity Fuzziness. 
For example, detailed modifiers in an instruction such as specifying attributes or exact counts can be removed, resulting in a more abstract entity description. 
As a result, the agent must infer the target object under reduced linguistic guidance.

\ding{183}\textbf{Procedural Compression.}
This category compresses action-related information by reducing procedural specificity while preserving executable intent. It can be formalized as
\begin{equation}
\tilde{\bm{a}} = T_a(\bm{a}), \quad \tilde{\bm{a}} \in \mathcal{A}_{\text{coarse}},
\end{equation}
where $\mathcal{A}_{\text{coarse}}$ denotes a space of higher-level or less specified actions. 	
This category includes the following subcategories: \emph{Action Replacement}, \emph{Action Generalization}, and \emph{Source Omission}.
Specifically, Action Replacement substitutes specific actions with semantically equivalent alternatives; Action Generalization compresses detailed actions into higher-level descriptions (\eg, ``\texttt{heat up the bread with the microwave}'' $\rightarrow$ ``\texttt{heat the bread}''); and Source Omission removes non-essential source information (\eg, removing ``\texttt{from the fridge}'' in ``\texttt{take the cup from the fridge}'').

\ding{184} \textbf{Constraint Relaxation.}
This category weakens or removes explicit constraints on task outcomes, particularly spatial relations, and can be expressed as
\begin{equation}
\tilde{\bm{c}} = T_c(\bm{c}), \quad \tilde{\bm{c}} \subseteq \bm{c},
\end{equation}
where $\tilde{c}$ defines a relaxed feasible constraint set. A typical subcategory is \emph{Spatial Constraint Relaxation}, which simplifies fine-grained placement conditions, such as transforming ``\texttt{place the cup on the counter next to the microwave}'' into ``\texttt{place the cup next to the microwave}''. By relaxing constraints, the feasible solution space is enlarged, thereby increasing uncertainty in spatial reasoning and target localization.

\ding{185} \textbf{Prior Knowledge Reduction.}
This category removes dependencies on implicit knowledge sources, including commonsense reasoning and prior observations. It can be formulated as
\begin{equation}
\tilde{\bm{p}} = T_p(\bm{p}), \quad \tilde{\bm{p}} \preceq \bm{p},
\end{equation}
where $p$ encodes perceptual memory and commonsense priors. This includes two subcategories: \emph{Commonsense Ignorance}, which eliminates cues that rely on external knowledge (\eg, ``\texttt{pick up the knife, cut the bread}'' becomes ``\texttt{cut the bread}''), and \emph{Visual Memory Removal}, which removes references to prior context (\eg, ``\texttt{put the cup back into the fridge}'' becomes ``\texttt{put the cup back}''). These transformations weaken auxiliary knowledge support and require the agent to rely more on current perception and intrinsic reasoning.

\medskip
\noindent
Overall, the proposed taxonomy defines a structured family of transformations $\{T_o, T_a, T_c, T_p\}$, each operating on a distinct semantic component. This provides a concrete instantiation of the abstract transformation operator $T$ introduced in \Eref{eq:transformation} and enables fine-grained analysis of how different information dimensions affect embodied agent decision-making.

\begin{figure}[!t]
\centering
\includegraphics[width=\linewidth,trim=280 260 300 290, clip]{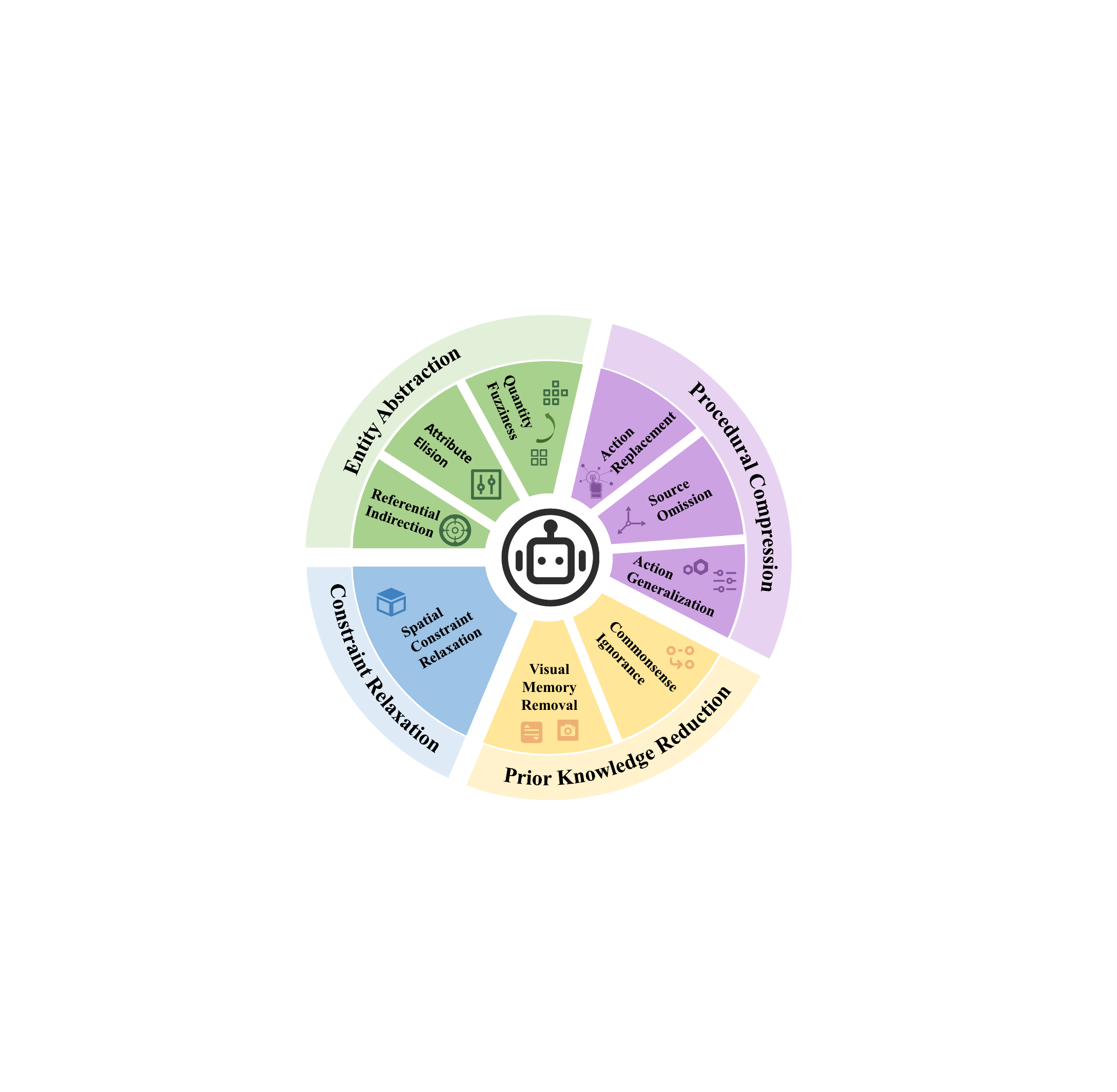}
\caption{Taxonomy of instruction in \bench, consisting of four high-level categories and nine fine-grained subtypes.}
\label{fig:vaguebench_taxonomy}
\end{figure}

\subsection{Dataset Construction}
To construct \bench\ with semantically consistent and diagnostically meaningful instruction pairs, we adopt a structured human–agent co-generation pipeline. 
The construction process consists of five sequential steps:

\ding{182} \textbf{Instruction sourcing.}
We collect high-quality, human-annotated instructions for long-horizon embodied tasks from Alfred \cite{shridhar2020alfred}. 
These instructions typically contain abundant auxiliary contextual information, making them suitable for controlled information reduction.

\ding{183} \textbf{Semantic redundancy estimation.}
Given an original instruction $\bm{I}$, an agent estimates its semantic redundancy to quantify how much contextual information can be removed without affecting the underlying task. 
This estimation considers multiple dimensions, including entity attributes, procedural specificity, constraint granularity, and reliance on prior knowledge.

\ding{184} \textbf{Human validation of estimation.}
To ensure that the estimated redundancy is semantically valid, human annotators inspect the candidate instructions. 
An instruction is retained only if its auxiliary components can be removed without altering the core task intent. 
This step verifies that the instruction indeed contains removable contextual information, and filters out cases where such information is indispensable for disambiguation.

\ding{185} \textbf{Taxonomy-guided transformation.}
For each validated instruction $\bm{I}$, we generate a context-reduced counterpart using the taxonomy-guided transformations $T(\cdot)$ defined in \Eref{eq:transformation}. 
These transformations systematically remove specific types of contextual cues.

\ding{186} \textbf{Semantic verification.}
Finally, we enforce the semantic-invariance condition $\mathcal{V}_{\text{human}}(\bm{I}, \tilde{\bm{I}})=1$ through human verification (detailed in \Sref{4.3}). 
Annotators confirm that the transformed instruction preserves the original task semantics and remains unambiguous and executable. 
This step ensures that the transformation introduces no semantic drift.

The final benchmark dataset comprises these meticulously verified $(I, \tilde{I})$ pairs, where instruction transformation serves as the sole controlled variable. Overall, this human-agent co-generation framework systematically isolates semantic information reduction as an independent factor. By varying only the availability of non-essential contextual cues while preserving task determinism, it provides a principled and controlled testbed for investigating the Ambient Context Illusion in embodied agents.

\subsection{Quanlity Control} \label{4.3}

To strictly enforce semantic invariance, we introduce a rigorous human verification process $\mathcal{V}_{human}(\bm{I}, \tilde{\bm{I}})$.  
This process ensures that the transformation removes only non-essential contextual cues rather than altering the core task semantics. 
Furthermore, it guarantees that the simplified instructions remain completely unambiguous within the given environment. 
Specifically, we adopt a \emph{triple-blind review protocol} for all instruction pairs. 
Three expert annotators independently evaluate each pair without access to the redundancy 
estimation scores, the applied transformation rules, or the decisions of other annotators. 
Annotators manually confirm that the underlying task objectives and the core operational intent remain strictly identical between the original instruction and its vague counterpart, and the omission of contextual details does not introduce multiple conflicting interpretations. 
An instruction pair is only retained for the final benchmark dataset if it receives unanimous approval from all three annotators. 
Through this independent triple-blind verification process, we robustly isolate semantic information reduction as the single controlled variable, ensuring the high quality and diagnostic validity of \bench.

\subsection{Dataset Statistics} \label{4.4}

% \begin{table}[t]

% \caption{Distribution across fuzziness categories, where percentages may exceed 100\% due to multi-label assignment.
% }
% \centering
% \label{tab:fuzziness_dist}
% \begin{tabular}{lcccc}
% \toprule
% \textbf{Metric} & \textbf{Entity} & \textbf{Procedural} & \textbf{Constraint} & \textbf{Prior} \\
% \midrule
% \# Samples   & 602   & 1034   & 415   & 204   \\
% Percentage   & 46.2\% & 79.4\% & 31.8\% & 15.7\% \\
% \bottomrule
% \end{tabular}

% \end{table}
\begin{figure}[t]
\centering
\includegraphics[width=\linewidth,trim=260 480 410 260, clip]{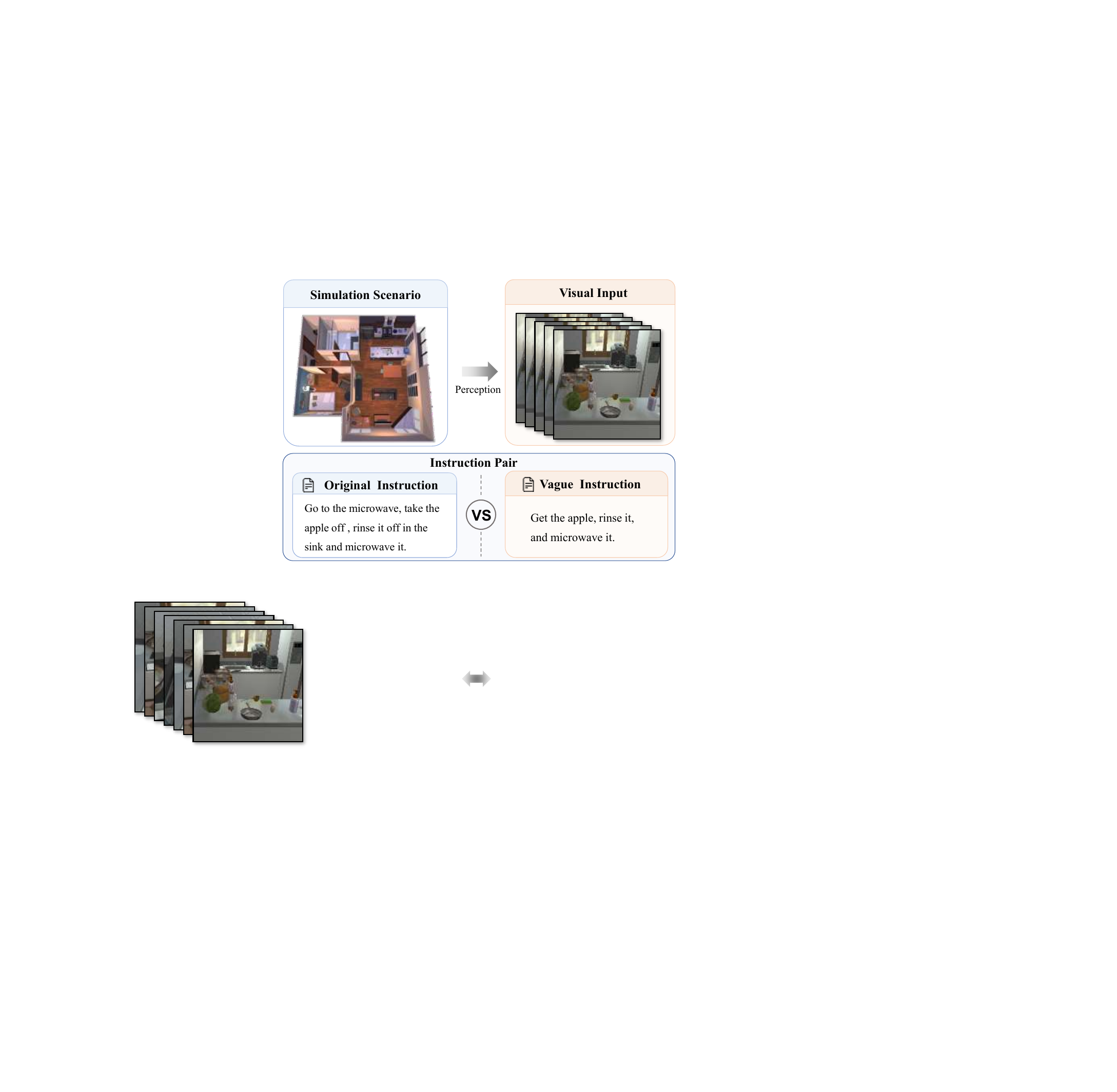}
\caption{Test example of \bench, consisting of an instruction pair and simulation scenario.}
\label{fig:test_example}
\end{figure}

\begin{figure}[t]
\centering
\includegraphics[width=\linewidth]{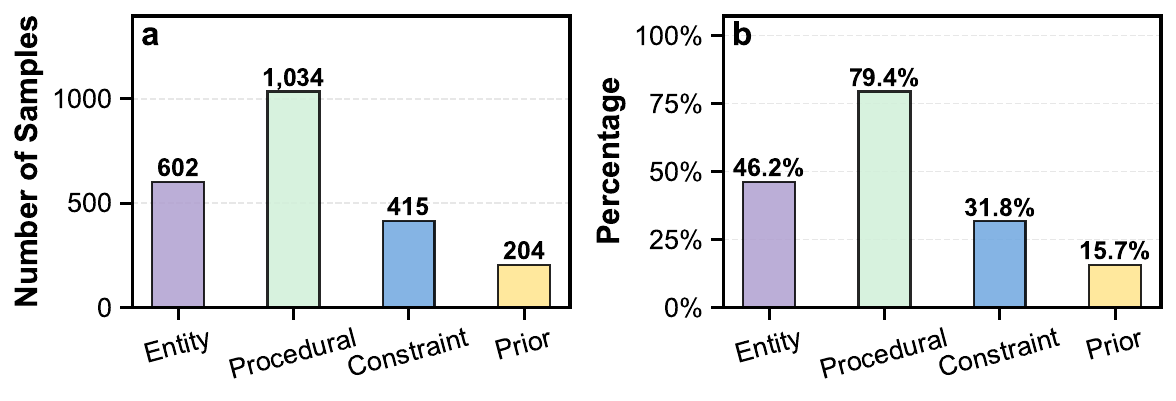}
\caption{Distribution across contextual cues categories.}
\label{fig:vaguebench_dist}
\end{figure}

\begin{table}[t]
\centering
\caption{Distribution across scene types.}
\label{tab:scene_dist}
\begin{tabular*}{\linewidth}{@{\extracolsep{\fill}}lcccc@{}}
\toprule
\textbf{Metric} & \textbf{Kitchen} & \textbf{Living Room} & \textbf{Bedroom} & \textbf{Bathroom} \\
\midrule
\# Samples   & 638   & 239   & 193   & 233   \\
Percentage   & 49.0\% & 18.3\% & 14.8\% & 17.9\% \\
\bottomrule
\end{tabular*}
\end{table}

\bench\ contains a total of 1,303 instruction pairs spanning diverse indoor environments and transformation approaches. For each test sample, we provide a task instruction pair and the corresponding simulation scenario, illustrated in \Fref{fig:test_example}. 

\textbf{Category distribution.}
The quantity of instruction pairs across the four main categories are demonstrated in \Fref{fig:vaguebench_dist}. 
Procedural Compression is the most prevalent transformation, suggesting that procedural details in natural language instructions are often redundant and can be simplified without affecting executability.
Since a single instruction may involve multiple types of redundant contextual cues, the categorization is inherently compositional, and each instruction can be associated with more than one category.
 
% In contrast, \emph{Prior Knowledge Reduction} occurs less frequently, indicating that dependencies on implicit knowledge tend to be more constrained and require careful handling to maintain task identifiability.

% From \Fref{fig:vaguebench_dist}, Procedural Compression is the most prevalent transformation, indicating that natural language instructions often contain redundant procedural details that can be removed without affecting executability.
% In contrast, Prior Knowledge Reduction appears less frequently, suggesting that dependencies on implicit knowledge are more constrained and must be handled carefully to preserve task identifiability.

\textbf{Scene distribution.}
\bench\ cover 4 common household task scenes, including kitchen, living room, bedroom, and bathroom scenes. As shown in \Tref{tab:scene_dist}, it maintains a balanced coverage across diverse embodied environments.
% We further report the distribution of instruction pairs across different scene types.
% Each instruction is associated with a single environment, making the scene distribution mutually exclusive and summing to 100\%.
% As shown in \Tref{tab:scene_dist}, the dataset maintains a balanced coverage across diverse embodied environments without over-representing any specific scenario.

Overall, \bench\ exhibits both compositional diversity in approaches to remove redundant contextual cues and balanced environmental coverage, providing a comprehensive testbed for analyzing how structured semantic information reduction affects embodied agent performance.

\section{Experiment}\label{5}

\begin{table*}[t]
\centering
\small
\setlength{\tabcolsep}{6pt}
\renewcommand{\arraystretch}{1.15}
\caption{Performance comparison under original and vague instructions. 
Each entry shows the base value followed by the absolute change. 
\textcolor{bettergreen}{Green} indicates increase and \textcolor{worsered}{red} indicates decrease.}
\label{tab:main_results_final}
\resizebox{\textwidth}{!}{
\begin{tabular}{lccccc}
\toprule
\textbf{Model} 
& \textbf{TSR \textcolor{red}{$\uparrow$}(\%)}
& \textbf{SCR \textcolor{red}{$\uparrow$}(\%)}
& \textbf{PE \textcolor{blue}{$\downarrow$}}
& \textbf{IAR \textcolor{blue}{$\downarrow$}(\%)}
& \textbf{EPR \textcolor{blue}{$\downarrow$}(\%)} \\
\midrule

Qwen2.5-VL-72B-Instruct 
& \hneg{43.51}{15.11}
& \hneg{49.88}{14.31}
& \lpos{22.41}{8.52}
& \lpos{22.35}{2.29}
& \lpos{10.67}{7.37} \\

Qwen2-VL-72B-Instruct
& \hneg{35.38}{5.53}
& \hneg{41.92}{6.89}
& \lpos{13.43}{2.30}
& \lneg{26.13}{0.21}
& \lpos{15.35}{6.29} \\

InternVL-3-78B
& \hneg{38.53}{6.91}
& \hneg{42.01}{6.87}
& \lpos{17.57}{1.00}
& \lpos{17.43}{1.95}
& \lpos{33.46}{5.83} \\

GPT-4o
& \hneg{\textbf{49.12}}{9.60}
& \hneg{\textbf{56.18}}{9.68}
& \lneg{\textbf{0.15}}{0.00}
& \lpos{\textbf{15.76}}{0.43}
& \lpos{10.28}{8.60} \\

\midrule

ProgPrompt
& \hneg{40.83}{5.60}
& \hneg{49.52}{5.83}
& \lneg{13.35}{2.22}
& \lneg{24.09}{1.12}
& \lpos{4.53}{5.14} \\

ReAct
& \hneg{31.01}{5.07}
& \hneg{38.16}{4.82}
& \lpos{140.75}{50.35}
& \lpos{38.67}{10.78}
& \lneg{\textbf{0.00}}{0.00} \\

\bottomrule
\end{tabular}
}
\end{table*}

\subsection{Experimental Setup}\label{5.1}

\textbf{Environment and benchmark.}
We conduct experiments in the AI2-THOR simulator \cite{kolve2017ai2}, a widely used platform for embodied agent research and use the evaluation framework in Embodiedbench \cite{yang2025embodiedbench}. Here, we consider diverse indoor environments, including kitchen, living room, bathroom, and bedroom scenes. We evaluate all methods on the proposed \bench, which contains a total of 1303 instruction pairs spanning diverse indoor environments. Each task is evaluated under both the original instruction $I$ and its corresponding modified version $\tilde{I}$.

\textbf{Target models.}
We evaluate 4 state-of-the-art VLMs: Qwen2.5-VL-72B-Instruct, Qwen2-VL-72B-Instruct, InternVL-3-72B, and GPT-4o.
These models represent diverse architectures and training paradigms, providing a comprehensive view of current capabilities in VLM-driven embodied agents.

\textbf{Agent workflows.}
Besides VLMs, we also evaluate our benchmark on different agent workflows. Here, we consider two representative paradigms: ProgPrompt \cite{singh2023progprompt}, which follows a program-driven execution scheme, and ReAct \cite{yao2022react}, which interleaves reasoning and action generation. This allows us to analyze how different interaction mechanisms affect robustness to ambient context illusions.

\textbf{Evaluation metrics.}
Following \cite{yang2025embodiedbench}, we conduct evaluations considering both effectiveness and reliability. For \emph{task performance}, we consider the Task Success Rate (TSR), defined as the proportion of tasks that are successfully completed, and the Subgoal Completion Rate (SCR), which measures partial progress by computing the fraction of completed subgoals in long-horizon tasks. For \emph{executablity}, we adopt three complementary metrics: Planner Output Error (PE), which measures the number of erroneous planner outputs, including malformed or execution-infeasible plans; Invalid Action Ratio (IAR), which measures the proportion of invalid actions during execution, reflecting the stability and correctness of the agent's policy; and Empty Plan Rate (EPR), which captures the frequency of empty or degenerate plans, indicating failure in high-level reasoning. For PE, IAR, and EPR, the higher values indicate better agent performance (\textcolor{blue}{$\downarrow$}); for TSR and SCR, the higher values indicate better agent performance (\textcolor{red}{$\uparrow$}).

%, and the Subgoal Completion Rate (SCR), which measures partial progress by computing the fraction of completed subgoals in long-horizon tasks.

%By comparing performance under original and vague instructions, we directly assess the sensitivity of embodied agents to the removal of non-essential contextual cues, and quantify the extent to which their behavior relies on implicit linguistic context rather than core task semantics.

% \begin{table*}[t]
% \caption{Performance comparison under original (base) and fuzzy instructions. 
% We report results in percentage (\%). Drop indicates the relative decrease: $(\text{base}-\text{fuzzy})/\text{base}$.}
% \centering
% \label{tab:main_results}
% \resizebox{\textwidth}{!}{%
% \begin{tabular}{lcccccc}
% \toprule
% \textbf{Model} & \textbf{SR (base)} & \textbf{SR (fuzzy)} & \textbf{Drop (\%)} & \textbf{SCR (base)} & \textbf{SCR (fuzzy)} & \textbf{Drop (\%)} \\
% \midrule

% Qwen2.5-VL-72B-Instruct & 43.51 & 28.40 & \textbf{34.73} & 61.38 & 43.77 & \textbf{28.69} \\
% Qwen2-VL-72B-Instruct   & 35.38 & 29.85 & 15.63 & 51.57 & 43.11 & 16.41 \\
% InternVL-3-72B          & 38.53 & 31.62 & 17.94 & 51.69 & 43.24 & 16.34 \\
% GPT-4o                  & \textbf{49.12} & \textbf{39.52} & 19.54 & \textbf{69.12} & \textbf{57.21} & 17.22 \\

% \midrule

% ProgPrompt              & 40.83 & 35.23 & 13.72 & 60.93 & 53.75 & 11.78 \\
% ReAct                  & 31.01 & 25.94 & 16.35 & 46.95 & 41.02 & 12.63 \\

% \bottomrule
% \end{tabular}
% }
% \end{table*}

\begin{table*}[t]
\centering
\small
\setlength{\tabcolsep}{6pt}
\renewcommand{\arraystretch}{1.15}
\caption{Performance comparison of different \bench\ taxonomy. 
Each entry shows the base value followed by the absolute change. 
\textcolor{bettergreen}{Green} indicates increase and \textcolor{worsered}{red} indicates decrease.}
\label{tab:vaguebench_results}
\resizebox{\textwidth}{!}{
\begin{tabular}{lccccc}
\toprule
\textbf{\bench\ Taxonomy} 
& \textbf{TSR \textcolor{red}{$\uparrow$}(\%)}
& \textbf{SCR \textcolor{red}{$\uparrow$}(\%)}
& \textbf{PE \textcolor{blue}{$\downarrow$}}
& \textbf{IAR \textcolor{blue}{$\downarrow$}(\%)}
& \textbf{EPR \textcolor{blue}{$\downarrow$}(\%)} \\
\midrule

Referential Indirection
& \hneg{33.33}{20.29}
& \hneg{41.67}{19.08}
& \lpos{28.99}{7.25}
& \lpos{27.68}{6.97}
& \lpos{7.25}{5.80} \\

Attribute Elision
& \hneg{35.63}{16.34}
& \hneg{43.03}{16.04}
& \lpos{20.47}{0.20}
& \lneg{25.48}{0.90}
& \lpos{10.63}{12.80} \\

Quantity Fuzziness
& \hneg{40.00}{24.00}
& \hneg{48.00}{25.00}
& \lpos{0.00}{32.00}
& \lpos{20.07}{6.81}
& \lpos{8.00}{20.00} \\

Action Replacement
& \hneg{40.93}{15.04}
& \hneg{48.01}{14.53}
& \lpos{28.98}{13.27}
& \lpos{22.91}{7.47}
& \lpos{9.51}{5.75} \\

Action Generalization
& \hneg{36.76}{22.06}
& \hneg{46.20}{22.55}
& \lpos{7.35}{42.65}
& \lpos{21.82}{12.46}
& \lpos{10.29}{7.35} \\

Source Omission
& \hneg{61.87}{14.20}
& \hneg{65.22}{13.00}
& \lpos{26.07}{5.06}
& \lpos{16.30}{2.39}
& \lpos{12.26}{7.59} \\

Spatial Constraint Relaxation
& \hneg{39.52}{18.55}
& \hneg{46.22}{17.39}
& \lpos{20.96}{19.76}
& \lpos{24.09}{3.90}
& \lpos{9.16}{4.58} \\

Commonsense Ignorance
& \hneg{39.13}{15.22}
& \hneg{45.52}{12.36}
& \lpos{31.52}{1.09}
& \lpos{22.94}{1.22}
& \lpos{14.13}{3.26} \\

Visual Memory Removal
& \hneg{40.00}{25.00}
& \hneg{47.50}{17.92}
& \lneg{95.00}{95.00}
& \lneg{21.98}{6.21}
& \lpos{5.00}{20.00} \\

\bottomrule
\end{tabular}
}
\end{table*}

\subsection{Main Evaluation Results}

Here, we first evaluate these VLM-based embodied agents on our \bench, and report their performance over the metrics. As shown in \Tref{tab:main_results_final}, we can identify:

%We removes non-essential contextual cues while preserving task semantics, to isolate the extent to which agents rely on auxiliary linguistic information. Overall, we observe consistent performance degradation across all models and metrics, indicating a systematic sensitivity to compressed instructions.

\ding{182} Across all models, we observe an consistent performance degradation on \bench, where both TSR and SCR exhibit noticeable declines when moving from original to vague instructions (\textbf{-7.97\%} and \textbf{-8.07\%} for TSR and SCR on average). For instance, Qwen2.5-VL-72B-Instruct shows a substantial drop of TSR (-15.11\%) and SCR (-14.31\%), while even the strongest model, GPT-4o, suffers a decrease of 9.60\% and 9.68\% on TSR and SCR, respectively. This consistent degradation suggests that, although vague instructions remain semantically complete, agents fail to fully recover the intended task structure without auxiliary contextual cues. This provides direct evidence that current embodied agents do not rely solely on core task semantics for decision-making. 

\ding{183} Redundant contextual cues removal also leads to a significant degradation in executablity (\textbf{+9.99}, \textbf{+2.35\%} and \textbf{+5.54\%} for PE, IAR and EPR on average). 
Beyond task performance, we observe consistent deterioration in executablity-related metrics. 
PE increases significantly for most models (\eg, +8.52 for Qwen2.5-VL-72B-Instruct), indicating more frequent planning failures. 
Similarly, IAR and EPR generally increase, suggesting reduced stability and a higher likelihood of degenerate behaviors. 
These results imply that auxiliary information not only affects high-level goal understanding but also propagates errors into low-level execution, which highlights a cascading failure effect from semantic ambiguity to action-level instability.

\ding{184} Workflow design influences robustness to auxiliary contextual cues.
Comparing other workflow-level methods, ProgPrompt demonstrates relatively smaller degradation in task performance (\textbf{-5.60\%} and \textbf{-5.83\%} for TSR and SCR) and even small improvements in certain executability metrics (\eg, reduced PE and IAR), indicating better structural robustness. 
In contrast, ReAct suffers from both lower absolute performance and significant increases in planner errors (+50.35) and invalid actions. 
A plausible explanation lies in the interaction between workflow design and environment feedback: in our setting, feedback signals are limited and largely binary, restricting the effectiveness of iterative reasoning in ReAct. 
Without rich state-level feedback, ReAct cannot effectively compensate for missing contextual information. 

Moreover, ReAct inherently invokes the planner at a much higher frequency than other workflows due to its iterative reasoning-acting paradigm. 
As a result, it accumulates substantially more planner calls over the course of an episode, which increases the exposure to potential planner failures. 
Even if the per-step error probability remains comparable, the higher number of planning steps amplifies the PE, leading to more frequent failures in practice. 
This suggests that structured, program-driven execution provides stronger inductive bias and better error control under compressed instructions, while iterative reasoning frameworks may be more vulnerable to error accumulation when contextual cues are reduced.

\textbf{Overall Summary.} 
Taken together, the consistent degradation across both effectiveness and executability metrics provides strong empirical support for the presence of Ambient Context Illusion. 
Even when instructions remain executable and semantically valid, removing redundant contextual cues leads to systematic performance loss and increased execution errors. 
This indicates that current embodied agents partially depend on implicit contextual signals, rather than fully grounding their behavior in core task semantics.

\subsection{Results on Different Vague Types}

\begin{figure}[!t]
    \centering
    \includegraphics[width=\linewidth]{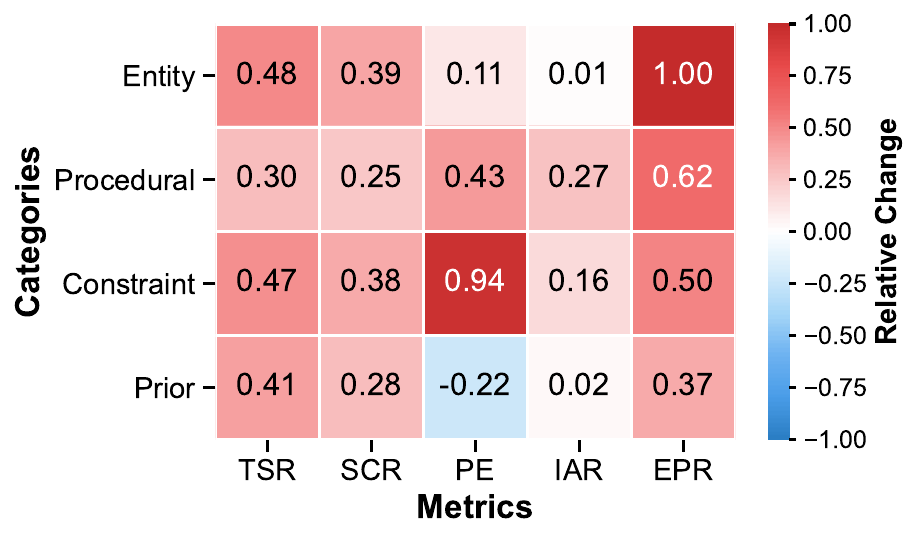}
    \caption{Diverging heatmap of relative performance changes across four categories and evaluation metrics. Positive values denote degradation under vague instructions, while negative values indicate improvement. Values are symmetrically clipped to $[-1.0,\,1.0]$ for better visual comparability.
    }
    \label{fig:heatmap}
\end{figure}

To better understand how different types of vague instructions affect embodied agents, we analyze four major categories and further examine representative sub-categories.

\textbf{Major Category Analysis.}
As shown in \Fref{fig:heatmap}, all four categories lead to consistent degradation in TSR and SCR, indicating that removing auxiliary linguistic information universally harms task performance. However, the underlying failure patterns differ significantly across categories.

\ding{182} Entity abstraction. For task performance, this type leads to the most severe relative degradation in TSR and SCR (-48.13\% and -38.91\%); for executability, we observe that PE slightly increases (+2.33), IAR remains largely unchanged (+0.32\%), while EPR rises sharply (+12.29 percentage points, corresponding to a relative increase of 121.31\%). This suggests that removing explicit object-related information primarily disrupts goal grounding, often preventing the agent from forming a valid plan rather than causing execution-level mistakes. In many cases, the agent fails at an early stage by selecting incorrect targets or producing underspecified goals, which subsequently propagate to downstream failures. We hypothesize that entity information serves as the primary anchor for mapping language to environment states; once attributes, references, or quantities are abstracted away, the agent struggles to resolve object identity, leading to ambiguity in goal specification and frequent degeneration into empty or incomplete plans \cite{ahn2022saycan,song2023llmplanner}.

% Constraint relaxation exhibits a comparable impact on task success but is characterized by a sharp increase in planner errors.
% This indicates that when spatial or logical constraints are underspecified, the agent can still infer the task goal but struggles to generate correct and coherent plans.

\ding{183} Constraint relaxation. For task performance, this type leads to a substantial relative drop in TSR and SCR (46.95\% and 37.62\%); for executability, however, it is characterized by a dramatic increase in PE (+19.76), accompanied by moderate increases in IAR (+3.90\%) and EPR (+4.58\%). This suggests that relaxing spatial or logical constraints does not prevent the agent from identifying the high-level task goal, but significantly impairs its ability to synthesize valid and coherent action sequences. In other words, goal recognition remains relatively intact, while plan construction becomes severely unstable. We hypothesize that constraint information functions as a structural prior for pruning the feasible action space; once underspecified, the agent must reason over a substantially enlarged and less constrained search space, leading to ambiguity and malformed plans \cite{fu2026beyond,shahnazari2026large}.

% Procedural compression results in relatively milder degradation in task success, yet introduces consistent increases across multiple error types.
% This pattern reflects a more distributed failure mode, affecting both planning and execution without a single dominant source of error.

\ding{184} Procedural compression. For task performance, this type results in a relatively milder degradation in TSR and SCR (-15.09 and -14.30 percentage points; relative drops of 29.55\% and 25.33\%); for executability, the effect is more distributed, where PE (+11.12, relatively +42.59\%), IAR (+5.27\%), and EPR (+6.77\%) all increase consistently. This pattern reflects a distributed failure mode: removing procedural details does not eliminate the agent’s understanding of the task goal, but weakens its ability to decompose the goal into a sequence of executable substeps. As a result, errors emerge across both planning and execution stages rather than concentrating in a single component. We hypothesize that procedural information provides an implicit hierarchical structure for long-horizon decision-making; when such structure is removed, the agent must infer missing subgoals, leading to compounded uncertainty throughout the pipeline \cite{choi2025reactree,liu2026any}.

\ding{185} Prior knowledge reduction. For task performance, this type leads to a moderate degradation in TSR and SCR (-16.18 and -12.91 percentage points; relative drops of 41.25\% and 28.24\%); for executability, it exhibits a distinct pattern where PE decreases (-8.33, relatively -22.08\%) while EPR increases (+4.90\%, relatively +37.04\%), with minimal change in IAR (+0.49\%). This suggests that removing commonsense or contextual priors simplifies the immediate planning process, reducing overt planning mistakes, but at the same time weakens the agent’s ability to infer implicit task requirements, resulting in more degenerate or incomplete plans. In this setting, the agent becomes more conservative: it produces fewer incorrect plans, yet also fails more often to produce sufficiently informative ones. We hypothesize that prior knowledge primarily supports implicit state inference and feasibility reasoning; without it, the agent avoids incorrect commitments but also loses the ability to complete tasks effectively \cite{chen2025re}.

\textbf{Key Sub-category Observations.}
While aggregated results provide a global picture, several sub-categories exhibit particularly strong or distinctive behaviors, as shown in \Tref{tab:vaguebench_results}.

% Referential indirection and quantity fuzziness lead to the most severe performance drops, with success rates decreasing by over 60\% relatively in some cases. 
% This highlights a fundamental weakness in object grounding: when explicit references (\eg, object names or counts) are removed, the agent often fails to correctly identify the target, leading to task failure.
\ding{182} Referential indirection and quantity fuzziness both lead to severe drops in TSR and SCR (-20.29\%, -19.08\% for TSR and -24.00\%, -25.00\% for SCR), but their error profiles differ. 
Referential indirection results in moderate and consistent increases across PE, IAR and EPR (+7.25, +6.97\%, +5.80\%), whereas quantity fuzziness causes a sharp surge in PE (+32.00). 
This suggests that beyond general grounding failure, quantity fuzziness additionally disrupts implicit task structure, leading to instability in plan construction rather than purely target misidentification. And comparing to other entity-relative information, reference and quantity contextual cues may have a more significant impact on object grounding \cite{zhu2025vgmamba}.

% Spatial constraint relaxation and action generalization show a different pattern. 
% In these settings, we observe a significant increase in planner output errors and invalid action ratios. 
% This indicates that the agent is still able to identify the correct goal, but struggles to generate precise and feasible action sequences, resulting in inefficient or unstable behavior.
\ding{183} Spatial constraint relaxation and action generalization further refine the coarse-grained observation on planning instability. 
While both increase PE, action generalization induces substantially larger degradation (+42.65 vs. +19.76). 
This indicates that removing procedural specificity is more disruptive than relaxing constraints, as it eliminates the internal action structure rather than merely expanding the feasible action space\cite{forouzandeh2025learning}.

% Source omission stands out as relatively less harmful compared to other types. 
% Although performance still degrades, the drop is smaller, suggesting that agents can partially rely on implicit spatial priors or environmental feedback to compensate for missing positional details.
\ding{184} Source omission stands out as relatively less harmful compared to other types. 
Although performance still degrades, the relative drop is smaller (-22.95\% for TSR and -19.93\% for SCR), and the increases in executability also remain small. 
This suggests agents can partially rely on implicit spatial priors or environmental observation to compensate for missing positional details\cite{zhang2026theory}.

\textbf{Analysis of Counterintuitive Trends.} Interestingly, we observe several counterintuitive cases. \ding{182} In attribute elision, the IAR slightly decreases (-0.90\%), which can be explained by the interaction between instruction simplification and the agent’s planning strategy.Removing detailed attributes or contextual cues can reduce over-specification, effectively shrinking the decision space. 
In such cases, the agent may generate simpler and more conservative action sequences, which can lead to fewer invalid actions even though the overall task success rate decreases. 
In other words, the agent becomes less expressive but also less prone to certain types of execution errors. \ding{183} In visual memory removal, PE drops significantly (-95.00). We assume that the absence of contextual information may suppress incorrect high-level reasoning paths. 
Without misleading contextual cues, the planner may avoid generating erroneous plans, resulting in lower planner output error rates. 
However, this does not necessarily translate to better task performance, as the agent may still fail to complete the task due to insufficient information.

\textbf{Overall Summary.}
The impact of auxiliary contextual cues is not uniform across categories or metrics. 
Entity-related information primarily affects goal grounding, procedural and constraint-related cues removal destabilize planning, and prior knowledge reduction influences high-level reasoning. 
The presence of counterintuitive metric trends further indicates that embodied agents exhibit complex dependencies on linguistic context, providing additional evidence for the existence of Ambient Context Illusion.

\begin{figure*}[!t]
    \centering
    \includegraphics[width=\linewidth,trim=80 390 80 330, clip]{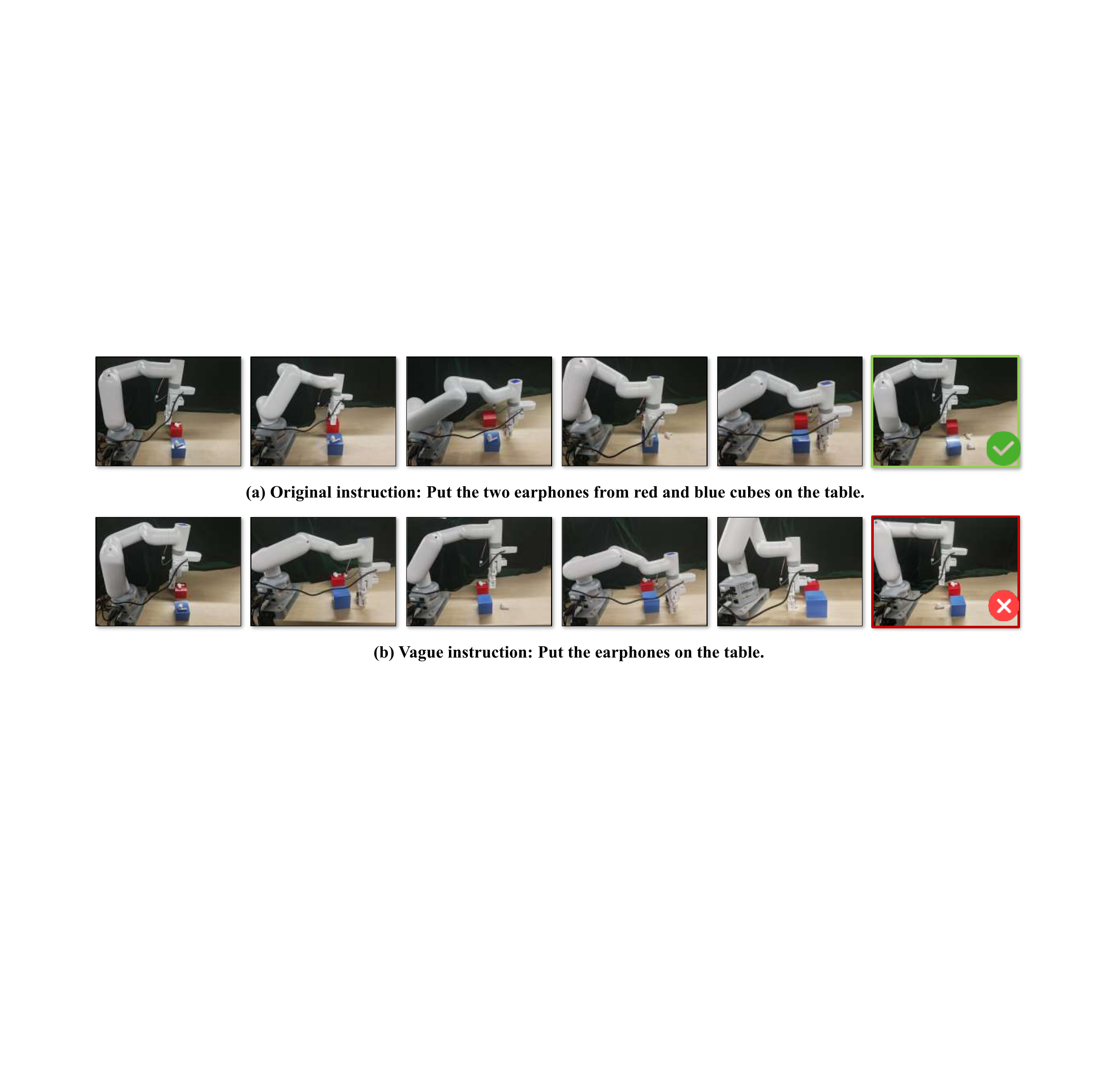}
    \caption{Real-world manipulation under original and vague instructions.}
    \label{fig:real_world_expr}
\end{figure*}

\subsection{Real-world Case Study} \label{real_world}

To verify whether Ambient Context Illusion persists beyond simulation, we conduct real-world case studies on a physical robotic platform.

\textbf{Experimental Setups.}
We employ a 6-DoF myCobot 280-Pi manipulator \cite{mycobot280pi2023} controlled by Qwen2.5-VL-Instruct-72B \cite{bai2025qwen25vltechnicalreport}, equipped with a adaptive gripper for object manipulation and an RGB camera for visual perception.
As illustrated in \Fref{fig:real_world_expr}, we construct paired instructions consisting of an original version with explicit contextual cues and a corresponding vague variant where such information is removed while preserving task semantics. Concretely, the original instruction (``\texttt{put the two earphones from red and blue cubes on the table}'') specifies both object identity and action source, whereas the vague version omits these auxiliary cues (``\texttt{put the earphones on the table}''). The task is evaluated across ten independent trials. Consistent with \Sref{5.1}, we adopt TSR as the evaluation metric, where a higher value indicates better task performance (\textcolor{red}{$\uparrow$}).

\textbf{Results Analysis.}
Under the original instruction, the agent correctly grounds the target objects and completes the task with coherent actions, achieving a TSR of \textbf{50\%}. In contrast, when the auxiliary cues are removed, the TSR drops to \textbf{10\%}, as the agent fails to reliably identify the intended objects, leading to incorrect selection and erroneous final states. The failure originates at the perception–grounding stage and propagates through execution.
This result demonstrates that removing auxiliary contextual cues leads to a significant degradation in task performance, even when the task remains well-defined. The effect is further amplified in real-world settings due to visual noise and object similarity.
Overall, the findings confirm that Ambient Context Illusion persists beyond simulation, revealing a fundamental limitation of current embodied agents and motivating the need for robustness under concise instructions.

\section{Mitigation Approach}

\begin{figure}[t]
    \centering
    \includegraphics[width=\linewidth]{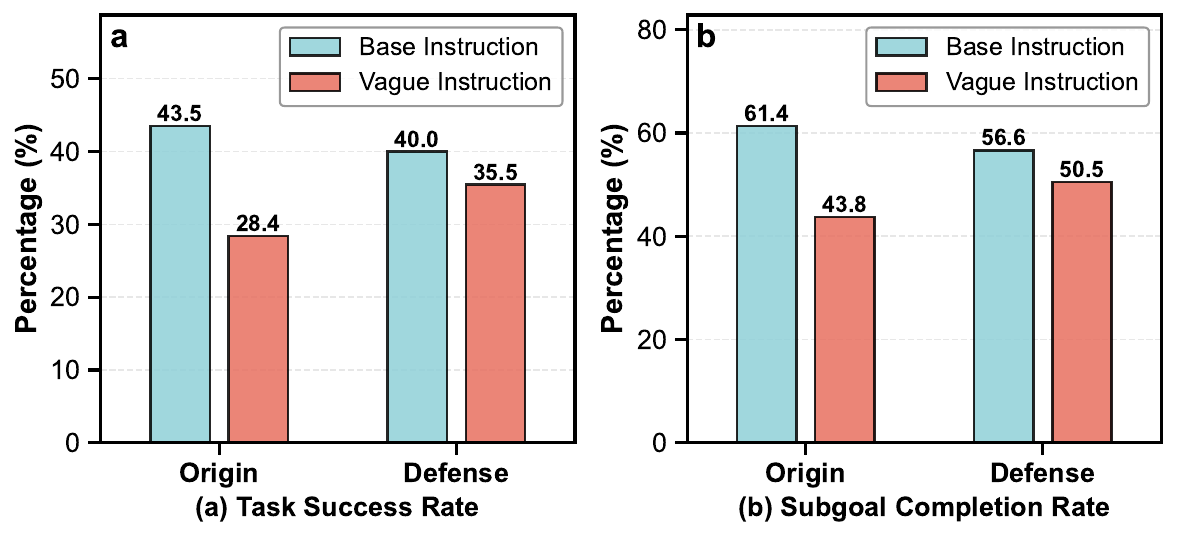}
    \caption{The mitigation effect of Context Suppression Constrain. }
    \label{fig:defense_results}
\end{figure}

% \begin{table*}[t]
% \centering
% \small
% \renewcommand{\arraystretch}{1.2}

% \caption{Effect of the proposed robustness constraint. 
% Drop denotes relative decrease $(\text{base}-\text{fuzzy})/\text{base}$.}
% \label{tab:defense_results}
% \resizebox{\textwidth}{!}{
% \begin{tabular}{lcccccc}
% \toprule
% \textbf{Setting} & \textbf{SR (base)} & \textbf{SR (fuzzy)} & \textbf{Drop (\%)} & \textbf{SCR (base)} & \textbf{SCR (fuzzy)} & \textbf{Drop (\%)} \\
% \midrule

% Origin  & 43.51 & 28.40 & 34.73 & 61.38 & 43.77 & 28.69 \\
% Defense & 39.98 & 35.46 & 11.29 & 56.65 & 50.50 & 10.86 \\

% \bottomrule
% \end{tabular}
% }
% \end{table*}

The results in \Sref{5} support for the presence of Ambient Context Illusion in both simulation and physical world.
Motivated by this, we introduce a lightweight mitigation strategy, termed \emph{Context Suppression Constraint} (CSC), which explicitly regulates how instructions are interpreted during decision-making, to mitigate the reliance on redundant contextual cues.

% The results in \Sref{5} reveal a consistent and substantial performance degradation of embodied agents when auxiliary information be removed. 
% Even when instructions remain semantically correct and unambiguous, agents lose a significant portion of their capability once contextual linguistic cues are removed. 
% This phenomenon, which we term Ambient Context Illusion, suggests that current agents partially rely on auxiliary language signals rather than fully grounding their behavior in core task semantics.

% To mitigate the performance degradation induced by redundant cues removal, we introduce an instruction-level robustness constraint that explicitly regulates how agents interpret natural language instructions during planning.

\subsection{Context Suppression Constrain}

\begin{figure*}[!t]
\centering
\includegraphics[width=\textwidth,trim=40 290 60 140, clip]{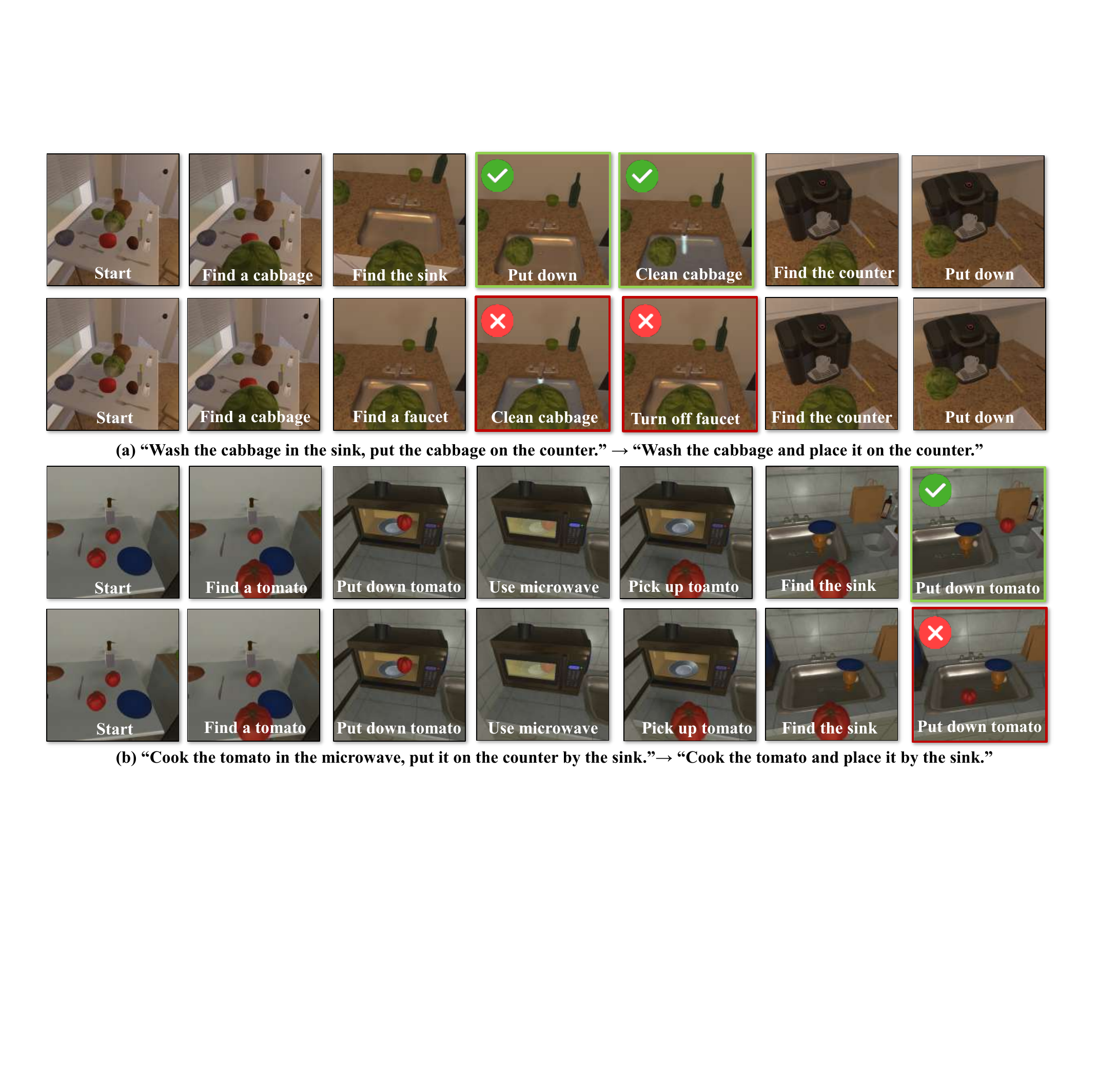}
\caption{Failure cases under original and vague instructions. 
The agent correctly executes tasks with explicit contextual cues but fails when such information is removed.}
\label{fig:failure_case}
\end{figure*}

Inspired by recent literature that advocates for runtime governance to constrain embodied execution \cite{qin2026harnessing} and highlights the necessity of preventing agents from making blind assumptions under ambiguity \cite{shen2025elba}, we propose to intervene directly at the inference stage. We introduce an instruction-level constraint $\bm{\phi}_{\mathrm{CSC}}$ that regularizes the semantic plan formation process.

Concretely, we reformulate the agent's decision process as
\begin{equation}
\bm{z}_t^* = f_{\Phi}(\bm{I}, \bm{o}_t, \bm{h}_t, \bm{\phi}_{\mathrm{CSC}}),
\label{eq:csc_policy}
\end{equation}
where $\bm{z}_t^*$ denotes the constrained semantic plan induced by the instruction under CSC. 
Specifically, CSC enforces three principles during instruction interpretation. 
\textbf{\ding{182} Goal-centric grounding}, where the agent prioritizes the target object and its desired final state as the core of the task; 
\textbf{\ding{183} no implicit assumption}, where missing attributes or constraints are not inferred beyond what is explicitly specified; 
and \textbf{\ding{184} context suppression}, where auxiliary information, such as redundant object attributes or prior states, is ignored unless required for task completion. 

From this perspective, CSC can be viewed as an inference-time regularization over plan formation. 
Instead of modifying the model parameters, it constrains how $\bm{I}$ is mapped to $\bm{z}_t^*$, encouraging a more stable and semantics-driven task execution plan that is less sensitive to contextual variations.

% Given an instruction $I$, the agent is guided to focus on essential task semantics rather than relying on implicit contextual cues. Specifically, the proposed constraint enforces three principles:
% \textbf{\ding{182} goal-centric grounding}, where the agent prioritizes the target object and its desired final state;
% \textbf{\ding{183} no implicit assumption}, where missing attributes or constraints are not inferred beyond what is explicitly specified;
% and \textbf{\ding{184} context suppression}, where non-essential information, such as auxiliary object attributes or prior states, is ignored unless required for task completion.

% This design is motivated by the observation that agents implicitly depend on auxiliary contextual information. In the absence of such cues, agents tend to compensate by introducing spurious assumptions, leading to incorrect object grounding, inefficient exploration, or unstable planning behaviors.

% By explicitly constraining the interpretation process, the proposed method effectively reduces the hypothesis space induced by incomplete linguistic signals, encouraging a direct mapping from instruction semantics to action sequences. From this perspective, the constraint can be viewed as a form of regularization on language understanding, enforcing reliance on task-relevant information.

% Moreover, this formulation directly targets the root cause of Ambient Context Illusion, the over-reliance on redundant contextual cues, thereby improving robustness without modifying model architectures or training procedures.

\subsection{Experimental Evaluation}

We evaluate CSC on Qwen2.5-VL-72B-Instruct under \bench, following the same setup as in \Sref{5}. 
As shown in \Fref{fig:defense_results}, CSC significantly improves performance under \bench. 
Without the constraint, the model exhibits a 34.73\% drop in SR and a 28.69\% drop in SCR, while the degradation decreases to 11.29\% and 10.86\% with CSC. 
This demonstrates that the agent becomes significantly less sensitive to redundant contextual cues with CSC.

We also observe a moderate performance decrease under original instructions, where SR drops from 43.51\% to 39.98\%. 
This suggests that CSC suppresses the model's ability to exploit auxiliary contextual signals, which may sometimes be beneficial when full information is available. 
Nevertheless, this trade-off leads to improved robustness and generalization under semantically minimal instructions.
Overall, CSC provides a simple yet effective mechanism for improving robustness in embodied agents. 
Its model-agnostic design makes it readily applicable to existing systems, and highlights the importance of controlling instruction interpretation for reliable embodied decision-making.

\section{Discussion}

\subsection{Failure Case Analysis}

\begin{figure}[t]
\centering
\includegraphics[width=\linewidth,trim=90 170 150 80, clip]{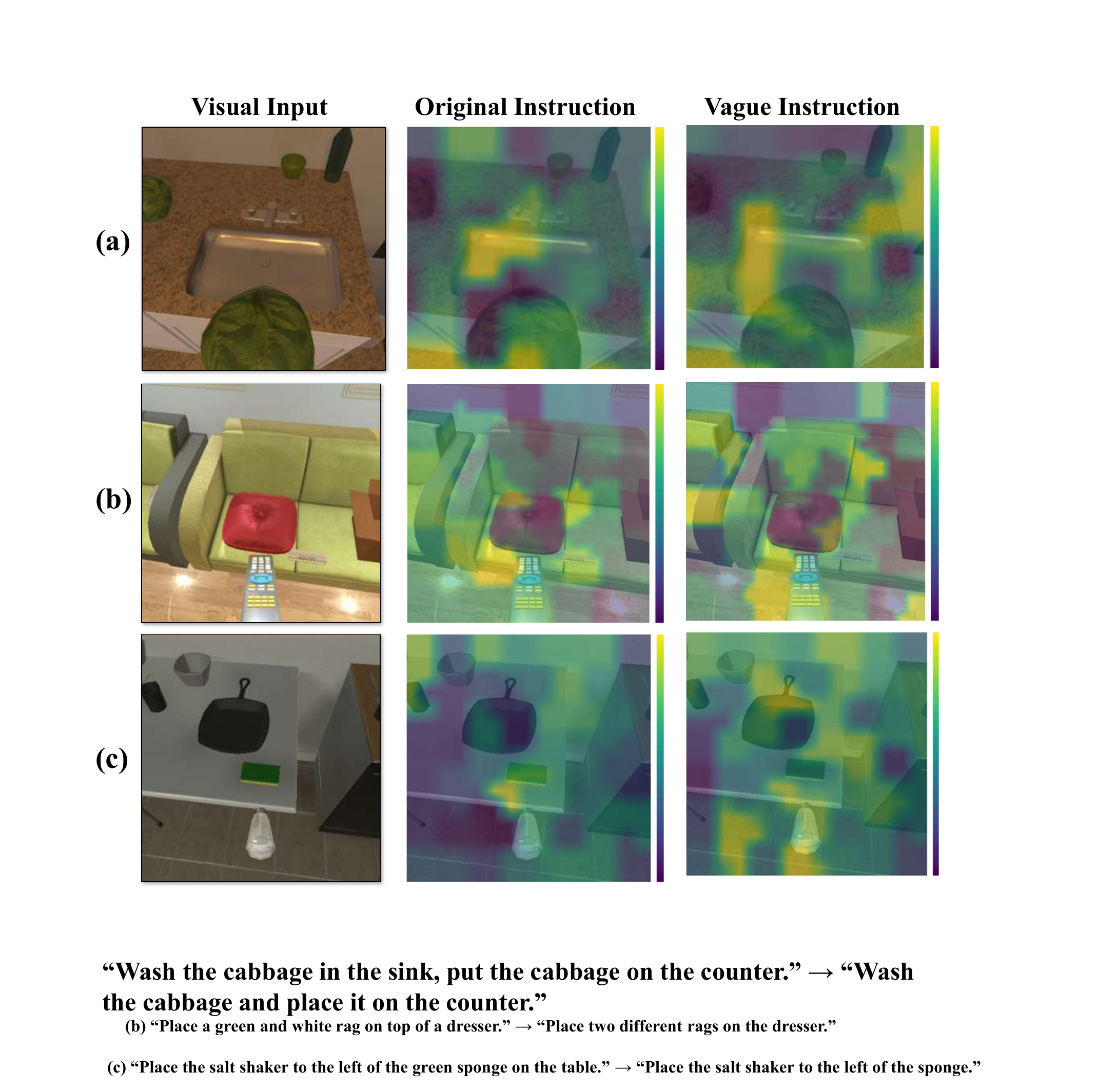}
\caption{Attribution maps that highlight the critical image regions driving model generation under both original and vague instruction.}
\label{fig:attribution_map}
\end{figure}

To better understand how vague instructions affects agents, we analyze representative failure cases.

% As shown in \Fref{fig:failure_case} (a), in the first example, the original instruction explicitly specifies both the interaction context and placement condition as ``\texttt{Wash the cabbage in the sink, put the cabbage on the counter.}'' Under this setting, the agent correctly locates the sink, performs the washing action, and places the cabbage at the intended location, following a coherent plan. However, when presenting the vague instruction ``\texttt{Wash the cabbage and place it on the counter}'', the agent deviates from the correct procedure. Instead of interacting with the sink, it performs the cleaning action using an alternative object (\eg, a faucet), resulting in an incorrect execution process despite reaching a superficially similar final placement.

% A similar pattern is observed in the second example, as shown in \Fref{fig:failure_case} (b). The original instruction ``\texttt{Cook the tomato in the microwave, put it on the counter by the sink}'' provides explicit information about both the required appliance and the target placement. In contrast, when the instruction is modified to ``\texttt{Cook the tomato and place it by the sink}'', the agent still performs the cooking action but fails to ground the placement constraint precisely. 
% As a result, the final placement deviates from the intended location, indicating a breakdown in spatial reasoning under reduced contextual guidance.

\textbf{Failure Case Illustration.} We first illustrate two failure cases. As shown in \Fref{fig:failure_case} (a), under the original instruction ``\texttt{Wash the cabbage in the sink, put the cabbage on the counter}'', the agent correctly executes a coherent plan by grounding the action to the sink. However, when simplified to ``\texttt{Wash the cabbage and place it on the counter}'', the agent substitutes the sink with a functionally similar object (\eg, faucet), leading to an incorrect interaction process despite a superficially correct outcome. This indicates that removing explicit interaction anchors weakens the mapping between actions and their canonical execution regions. As shown in \Fref{fig:failure_case} (b), for the instruction ``\texttt{Cook the tomato in the microwave, put it on the counter by the sink}'', the agent accurately follows both the appliance and spatial constraint. In contrast, the vague version ``\texttt{Cook the tomato and place it by the sink}'' results in imprecise placement. While the high-level task structure is preserved, the agent fails to bind the relational constraint to a specific executable location, revealing a breakdown in spatial reasoning under reduced context.

% These cases reveal that removing seemingly auxiliary contextual details (such as specific interaction locations or reference anchors), can lead to systematic deviations in both action selection and placement decisions. 
% Although the compressed instructions remain semantically valid, the agent fails to fully align its behavior with the intended task specification. Notably, the failures do not always manifest as complete breakdowns, but often as subtle deviations in execution or final state. This suggests that contextual cues can impair different stages of the decision-making process, including interaction grounding and spatial reasoning, even when the overall task structure is preserved.

\textbf{Analysis.}
In case (a), the failure stems from incorrect grounding. Without the explicit cue ``\texttt{in the sink}'', the agent fails to map the verb ``\texttt{wash}'' to its canonical interaction region. Instead, it selects objects with partial functional similarity (\eg, a faucet), suggesting a reliance on surface-level correlations rather than true action understanding \cite{nguyen2025causalplan}.
Conversely, case (b) highlights spatial constraint failures. When the cue ``\texttt{by the sink}'' is weakened, the agent maintains the high-level task sequence (\ie, cooking then placing) but fails to target a precise location. This yields goal-consistent yet spatially inaccurate outcomes, demonstrating that embodied agents rely heavily on explicit cues to constrain their solution space \cite{liu2025can}. Ultimately, both cases stem from an expanded, weakly constrained hypothesis space caused by Ambient Context Illusion. Omitting auxiliary context forces agents to implicitly infer constraints, broadening task interpretations. Without strong structural priors, they select superficially valid but semantically misaligned solutions. Agents preserve the overall task skeleton but deviate in critical details, highlighting that the issue lies not in complete task misunderstanding, but in imprecise core semantic grounding.

\textbf{Visual Interpretation.}
To further support this, we apply the EAGLE framework \cite{chen2025mllms} to explain the agent's autoregressive token generation during key decision steps, comparing the model's visual reliance on task-relevant regions under original and vague instructions. Importantly, these attribution maps visualize the causal drivers of the model's predictions, where high-intensity regions denote the visual features most crucial to the agent's decisions. As shown in \Fref{fig:attribution_map}, under explicit instructions, the agents are sharply concentrated on the correct interaction anchors (\eg, sink and cabbage). Conversely, under the vague instruction, the model's reliance becomes more diffuse or shifts toward visually salient but task-irrelevant objects like countertops and walls. This behavior is consistent with the hypothesis that the agent relies on surface-level correlations provided by auxiliary contextual cues, rather than really understanding the core task semantics.

\subsection{Effects of Model Scale}

\begin{figure}[t]
\centering
\includegraphics[width=\linewidth]{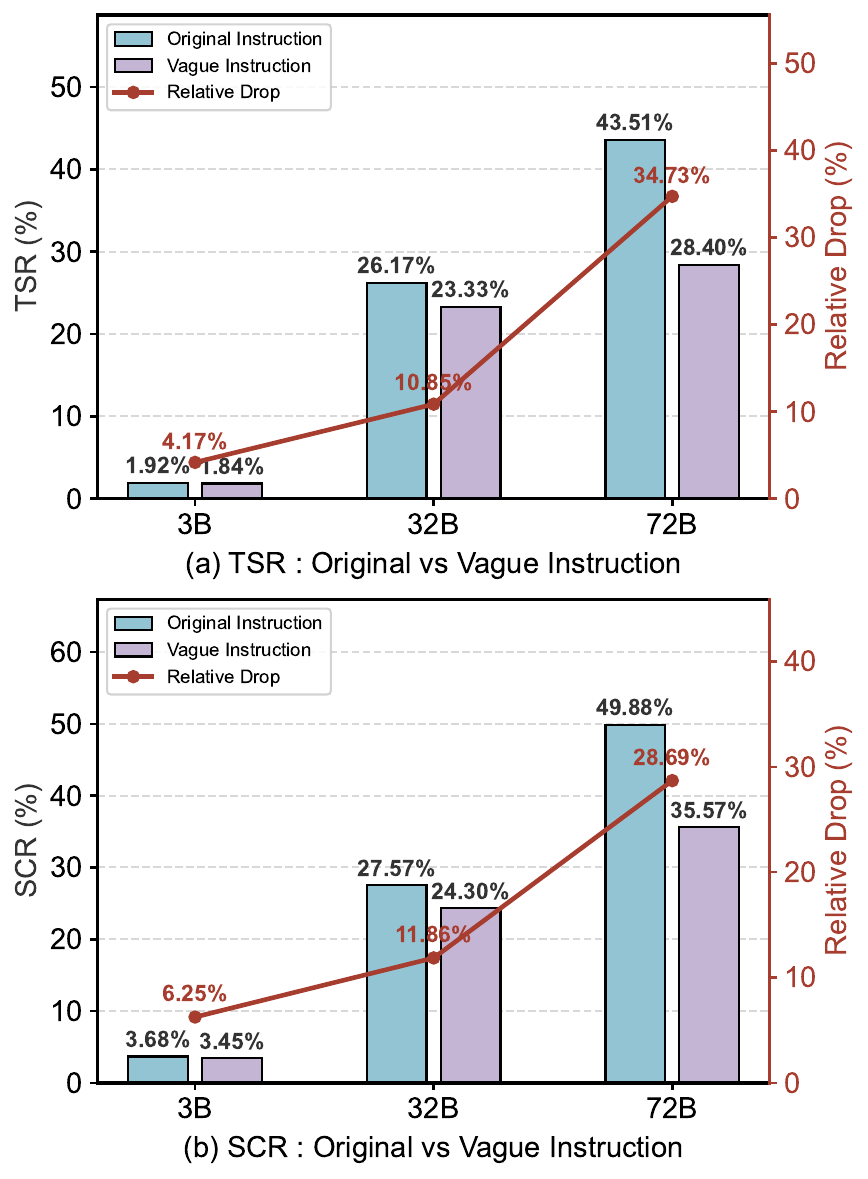}
\caption{Task performance across different scales of Qwen2.5-VL under original and vague instructions. Bars correspond to absolute TSR/SCR scores (left y-axis, black), whereas the red line indicates the relative performance drop (\%) (right y-axis, red).}
\label{fig:model_scale}
\end{figure}

A natural question is whether Ambient Context Illusion can be alleviated simply by scaling up model size. 
To examine this possibility, we further conduct a controlled scaling analysis within the same model family (Qwen2.5-VL \cite{bai2025qwen25vltechnicalreport}), comparing models of different sizes on \bench\ under the same evaluation protocol. Specifically, we measure the performance degradation from original instructions to vague instructions using TSR and SCR, and focus on the relative drop rather than absolute performance.

As shown in \Fref{fig:model_scale}, we observe a counterintuitive trend: as model scale increases, the relative degradation becomes even more severe. Specifically, the relative TSR drop grows monotonically from 4.2\% for the 3B model (from 1.9\% to 1.8\%), to 10.9\% for the 32B model (from 26.2\% to 23.3\%), and further to 34.7\% for the 72B model (from 43.5\% to 28.4\%). Although larger models achieve substantially stronger absolute performance under the original instruction setting, their performance declines much more sharply once auxiliary contextual cues are removed.
In other words, scaling does not improve robustness to context reduction; instead, it amplifies sensitivity to redundant contextual information. This finding suggests that Ambient Context Illusion is not a simple consequence of limited model capacity. If the phenomenon were primarily caused by insufficient scale, larger models would be expected to exhibit smaller relative degradation. However, our results show the opposite: the degradation magnitude increases by over 8× from 3B to 72B.
This indicates that larger models may rely more strongly on rich contextual correlations learned during pretraining and instruction tuning, and thus become more vulnerable when such auxiliary cues are suppressed. We hypothesize that model scaling strengthens the ability to exploit contextual regularities, but does not necessarily improve goal-centric semantic grounding \cite{wang2026refact,zhao2026large}. As a result, larger models may perform better when instructions are richly specified, yet fail more noticeably when only the core task semantics are preserved.

Overall, this result provides additional support for our main conclusion: improving embodied agent robustness requires mechanisms that explicitly promote understanding on core task semantics, rather than expecting model scale itself to resolve the problem.

\subsection{Compounding Effect of Multiple Vague Instruction Types}

\begin{figure}[t]
\centering
\includegraphics[width=\linewidth]{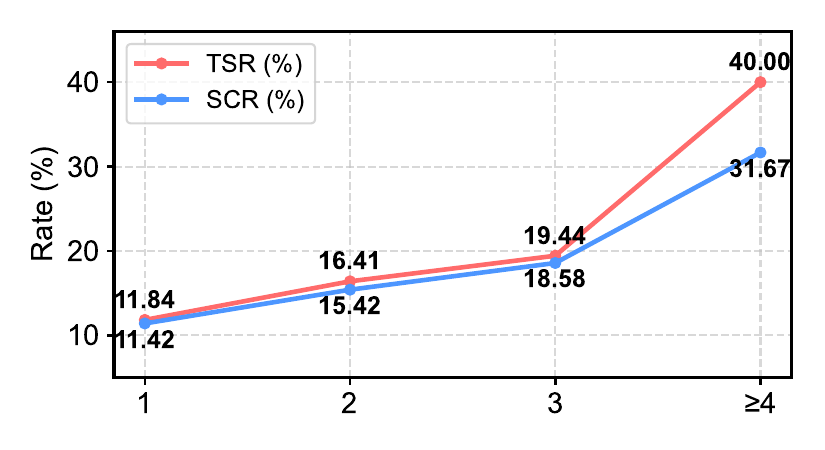}
\caption{Performance degradation under increasing numbers of vague instruction types.}
\label{fig:type_number}
\end{figure}

While previous analysis isolates the effect of individual transformation types, real-world instructions are typically subject to compositional simplification, where multiple semantic dimensions are simultaneously altered. As discussed in \Sref{4.4}, \bench\ is inherently multi-label, allowing a single instruction to involve multiple transformation subcategories. This raises a critical question: whether performance degradation under multiple transformations follows a simple additive pattern, or exhibits more complex interaction effects. 
To answer this, we group vague instructions by the number of applied transformation types (1, 2, 3, and $\geq 4$), and measure the corresponding degradation in TSR and SCR. 

As shown in \Fref{fig:type_number}, performance degradation increases monotonically with the number of transformations, growing from \textbf{11.84\%} under a single transformation to \textbf{40.00\%} when four or more are applied. However, this trend is distinctly non-linear: the degradation from one to three transformations remains relatively moderate, whereas the transition to four or more results in a sharp and disproportionate collapse.
If each transformation contributed independently, the total degradation would approximate the sum of individual effects (\ie, an additive regime). Instead, the observed acceleration indicates a compositional effect, where multiple transformations interact in a mutually amplifying manner.
A plausible explanation lies in the progressive erosion of auxiliary contextual cues. Individual transformations remove specific redundancies (\eg, entity attributes, constraints, or procedural hints), which the agent can partially compensate for. However, when multiple such cues are simultaneously suppressed, the remaining signal becomes insufficient to support stable goal-centric semantic grounding. At this stage, the agent no longer exhibits graceful degradation, but rather a systemic breakdown in interpretation and planning.

Collectively, these results suggest that Ambient Context Illusion does not arise from isolated sensitivity to individual information omissions, but from entangled dependencies on multiple contextual cues, whose joint removal triggers a non-linear collapse in task performance.

\subsection{Sim-to-Real Gap}

Building on the single-task case study in \Sref{real_world}, this section systematically analyzes the sim-to-real gap of the Ambient Context Illusion. 
A key observation is that performance degradation from context reduction is drastically amplified in the real world. While simulation shows a moderate task success rate (TSR) drop of 15.11\% (\Tref{tab:main_results_final}), real-world experiments in \Sref{real_world} reveal a 40\% drop (from 50\% to 10\%). 
To further validate that this discrepancy is not incidental but systematic, we extend the real-world evaluation to a more comprehensive setting. Specifically, for each transformation subtype in our taxonomy, we design three instruction pairs, resulting in a total of 54 instructions. Each instruction is executed in five independent trials, yielding 270 real-world experiments in total. This controlled and repeated evaluation enables a more statistically reliable analysis of the sim-to-real gap across different types of contextual reduction. 
The experimental results demonstrate that TSR drops by 31.85\% (from 77.78\% to 45.93\%). 
It is worth noting that, due to the inherent constraints of physical environments, the task settings in the real world are significantly simpler than those in simulation, leading to a relatively higher baseline success rate. 
However, despite this reduced task complexity, the magnitude of performance degradation remains substantially larger than that observed in the simulator. This indicates that Ambient Context Illusion is further amplified in real-world conditions.

This amplification may stems from three compounding factors:
\ding{182} Visual Perception Noise: Unlike clean simulators, real-world perception suffers from sensor noise, lighting variations, and visual ambiguity, making visual grounding highly unreliable when contextual cues are missing.
\ding{183} Execution Noise: Deterministic simulated actions contrast sharply with real-world kinematic and control noise. This physical execution noise compounds the semantic uncertainty caused by vague instructions, resulting in cascading errors.
\ding{184} Stochastic Dynamics: Simulators are largely deterministic, whereas real environments are inherently unpredictable. Without contextual constraints, the expanded action space interacts with this stochasticity, significantly increasing failure modes.

\textbf{Summary.} The Ambient Context Illusion is structurally amplified in real-world deployments. The sim-to-real gap magnifies existing failures through perception, execution, and environmental uncertainties. Consequently, robustness to context reduction is a prerequisite for reliable real-world embodied agents, as simulation results often significantly overestimate physical performance.
\section{Conclusion And Future Work}
In this work, we investigate the robustness of embodied agents under the removal of semantically-preserving contextual cues and identify a novel phenomenon termed Ambient Context Illusion.
We show that eliminating non-essential cues leads to significant performance degradation, revealing a systematic reliance on auxiliary linguistic information.
To support controlled analysis, we construct \bench\ and conduct extensive evaluations across models, workflows, and real-world settings.
We further propose a lightweight, model-agnostic constraint CSC that mitigates this issue via goal-centric interpretation.
Overall, our results highlight that robustness in embodied agents requires stronger semantic grounding rather than only increased model capacity.
% In this work, we study the robustness of embodied agents under semantically-preserving redundant contextual cues removal and identify a novel phenomenon, termed Ambient Context Illusion.
% We show that even when task semantics remain unchanged, removing non-essential contextual cues leads to significant performance degradation, revealing a systematic reliance on auxiliary linguistic information.
% To enable controlled analysis, we construct \bench, a benchmark that categorizes redundant contextual cues across multiple dimensions and supports fine-grained evaluation.
% Extensive experiments across diverse models, workflows, and real-world settings demonstrate that current agents fail to consistently ground their behavior in core task semantics, and instead depend on redundant contextual signals.
% We further propose a lightweight, model-agnostic robustness constraint that mitigates this issue by enforcing goal-centric interpretation.
% Overall, our findings suggest that improving robustness in embodied agents fundamentally requires stronger semantic grounding rather than increased model capacity.

\textbf{Limitation.}
We would like to explore the following directions in future work: \ding{182} extending our evaluation to more diverse environments and more complex real-world robotic systems to improve generalization; and \ding{183} developing more principled, learning-based approaches beyond prompt-level constraints to enhance intrinsic semantic grounding under complex scenarios.

% Our study is primarily conducted in indoor environments and on a limited set of current VLM-driven embodied agents, and the generality of our findings to broader settings requires further investigation.
% In addition, failures caused by incorrect grounding or instruction misinterpretation may lead to unintended behaviors in real-world deployment.
% These observations highlight the importance of developing more robust and reliable embodied systems under realistic human instructions.
\section*{Statements and Declarations}

\textbf{Funding.} This work was supported by the New Generation Artificial Intelligence-National Science and Technology Major Project (2025ZD0123201) and National Natural Science Foundation of China(62576019, 62476018).

\textbf{Conflict of interest.} The authors have no relevant financial or non-financial interests to disclose.

\textbf{Consent for Publication.} Informed consent. 

\textbf{Author Contribution.}
Conceptualization: Yang Qu, Qiang Fu; Methodology: Yang Qu, Qiang Fu, Zonghao Ying; Formal analysis and investigation: Yang Qu, Jiayi Zhao, Yaoyuan Zhang; Writing - original draft preparation: Yang Qu, Qiang Fu, Zonghao Ying, Jiayi Zhao, Yaoyuan Zhang; Writing - review and editing: Jinyang Guo, Jian Yang, Aishan Liu, Yingxun Wang, Xianglong Liu; Funding acquisition: Aishan Liu, Xianglong Liu; Resources: Aishan Liu, Yingxun Wang, Xianglong Liu. 

\textbf{Consent to Participate.} Informed consent. 

\textbf{Ethical Statement and Broader Impact.} This study uses no human subjects or sensitive data. All experiments are conducted in simulation or controlled settings. \bench\ is intended for research to improve embodied agent performance, with no foreseeable ethical or societal concerns.

\textbf{Data Availability Statement.} 
The \bench\ codebase is hosted and version-tracked on GitHub, and is permanently available at \url{https://github.com/stardreaming/VagueBench}. The \bench\ dataset is hosted and version-tracked on Hugging Face and is permanently accessible at \url{https://huggingface.co/datasets/stardreaming/VagueBench}. We encourage the community to provide feedback and engage in scrutinizing the benchmark.

%%===========================================================================================%%
%% If you are submitting to one of the Nature Portfolio journals, using the eJP submission   %%
%% system, please include the references within the manuscript file itself. You may do this  %%
%% by copying the reference list from your .bbl file, paste it into the main manuscript .tex %%
%% file, and delete the associated \verb+\bibliography+ commands.                            %%
%%===========================================================================================%%

\bibliographystyle{unsrt}
\bibliography{main}% common bib file
%% if required, the content of .bbl file can be included here once bbl is generated
%%\input sn-article.bbl

% %%%%%%%%%%           APPENDIX           %%%%%%%%%%%%
% \clearpage
% \appendix
% \section*{Appendix}

% \vspace{1em}
% \hrule
% \vspace{1em}
% \input{Sections_supp/sec_1}
% \input{Sections_supp/sec_2}
% \input{Sections_supp/sec_3}
% \input{Sections_supp/sec_4}
% \input{Sections_supp/sec_5}
% % \input{Sections_supp/sec_6}
% \input{Sections_supp/sec_7}
% %%%%%%%%%%           APPENDIX           %%%%%%%%%%%%

\end{document}